\documentclass[10pt,twocolumn,letterpaper]{article}

\usepackage[pagenumbers]{cvpr} 

\makeatletter
\@namedef{ver@everyshi.sty}{}
\makeatother
\usepackage{graphicx}
\usepackage{amsmath, bm}
\usepackage{amssymb}
\usepackage{mathtools}
\usepackage{amsthm}
\usepackage{booktabs}
\usepackage{hhline}
\usepackage{multirow, makecell}

\usepackage{hyperref}
\hypersetup{pagebackref,breaklinks,colorlinks}
\usepackage[dvipsnames, table]{xcolor}
\usepackage{algorithm}
\usepackage{algpseudocodex}

\usepackage[capitalize]{cleveref}
\crefname{section}{Sec.}{Secs.}
\Crefname{section}{Section}{Sections}
\Crefname{table}{Table}{Tables}
\crefname{table}{Tab.}{Tabs.}
\newcommand{\bftab}{\fontseries{b}\selectfont}

\begin{document}

\title{Confidence-aware Personalized Federated Learning via \\Variational Expectation Maximization}

\author{Junyi Zhu$^*$\\ 
ESAT-PSI, KU Leuven\\
{\tt\small junyi.zhu@esat.kuleuven.be}
\and Xingchen Ma$^{*\dagger}$\\
Amazon Web Services\\
{\tt\small xgchenma@amazon.de}
\and Matthew B. Blaschko\\
ESAT-PSI, KU Leuven\\
{\tt\small matthew.blaschko@esat.kuleuven.be}
}

\maketitle
\def\thefootnote{*}\footnotetext{Authors with equal contribution.}
\def\thefootnote{$\dagger$}\footnotetext{Work was done at KU Leuven prior to joining Amazon.}
\def\thefootnote{\arabic{footnote}}
\begin{abstract}
Federated Learning (FL) is a distributed learning scheme to train a shared model across clients. One common and fundamental challenge in FL is that the sets of data across clients could be non-identically distributed and have different sizes. Personalized Federated Learning (PFL) attempts to solve this challenge via locally adapted models. In this work, we present a novel framework for PFL based on hierarchical Bayesian modeling and variational inference. A global model is introduced as a latent variable to augment the joint distribution of clients' parameters and capture the common trends of different clients, optimization is derived based on the principle of maximizing the marginal likelihood and conducted using variational expectation maximization. Our algorithm gives rise to a closed-form estimation of a confidence value which comprises the uncertainty of clients' parameters and local model deviations from the global model. The confidence value is used to weigh clients' parameters in the aggregation stage and adjust the regularization effect of the global model. We evaluate our method through extensive empirical studies on multiple datasets. Experimental results show that our approach obtains competitive results under mild heterogeneous circumstances while significantly outperforming state-of-the-art PFL frameworks in highly heterogeneous settings. Our code is available at~\url{https://github.com/JunyiZhu-AI/confidence_aware_PFL}.
\end{abstract}

\section{Introduction}
\label{sec:intro}
Federated learning (FL) is a distributed learning framework, in which clients optimize a shared model with their local data and send back parameters after training, and a central server aggregates locally updated models to obtain a global model that it re-distributes to clients~\cite{pmlr-v54-mcmahan17a}.
FL is expected to address privacy concerns 
and to exploit the computational resources of a large number of edge devices. Despite these strengths, there are several challenges in the application of FL. One of them is the statistical heterogeneity of client data sets since in practice clients' data correlate with local environments and deviate from each other~\cite{fl_survey,Li2020On, MLSYS2020_38af8613}. The most common types of heterogeneity are defined as:

\textbf{Label distribution skew.}
Let $J$ be the number of clients and the data distribution of client $j$ be $P_j(x,y)$ and rewrite it as $P_j(x | y) P_j(y)$, two kinds of non-identical scenarios can be identified. One of them is \textit{label distribution skew}, that is, the label distributions $\{P_j(y)\}_{j=1}^J$ are varying in different clients but the conditional generating distributions $\{P_j(x|y)\}_{j=1}^J$ are assumed to be the same. This could happen when certain types of data are underrepresented in the local environment.

\textbf{Label concept drift.}
Another common type of non-IID scenario is \textit{label concept drift}, in which the label distributions $\{P_j(y)\}_{j=1}^J$ are the same but the conditional generating distributions $\{P_j(x|y)\}_{j=1}^J$ are different across different clients. This could happen when features of the same type of data differ across clients and correlates with their environments, e.g. the Labrador Retriever (most popular dog in the United States) and the Border Collie (most popular dog in Europe) look different, thus the dog pictures taken by the clients in these two areas contain \textit{label concept drift}.

\textbf{Data quantity disparity.}
Additionally, clients may possess different amounts of data. Such \textit{data quantity disparity} can lead to inconsistent uncertainties of the locally updated models and heterogeneity in the number of local updates. In practice, 
the amount of data could span a large range across clients, for example large hospitals usually have many more medical records than clinics. In particular, data quantity distributions often exhibit that large datasets are concentrated in a few locations, whereas a large amount of data is scattered across many locations with small dataset sizes~\cite{Thessen2012,heidorn2008}.

It has been proven that if federated averaging (\verb "FedAvg" \cite{pmlr-v54-mcmahan17a}) is applied, the aforementioned heterogeneity will slow down the convergence of the global model and in some cases leads to arbitrary deviation from the optimum~\cite{wang_obj_inconsistency,Li2020On}. Several works have been proposed to alleviate this problem~\cite{MLSYS2020_38af8613, wang_obj_inconsistency, chen2021fedbe}.
Another stream of work is personalized federated learning (PFL) \cite{smith_neurips_mocha, fallah_perfedavg, Collins_icml_fedrep, dinh_pfedme, zhang_pfedbayes}, which trains multiple local models instead of a single global model.
Most PFL frameworks still construct and optimize a shared model which serves as a constraint or initialization for the personalization of local models. In the theoretical analysis of most existing PFL frameworks, the un-weighted average of clients' model parameters is used as the shared model, however, to obtain better empirical results, a weighted average is usually used and the weights depend on the local data size. This inconsistency indicates a more principled method is needed to optimize the shared model.


\begin{figure}[t!]
    \centering
    \includegraphics[width=\linewidth]{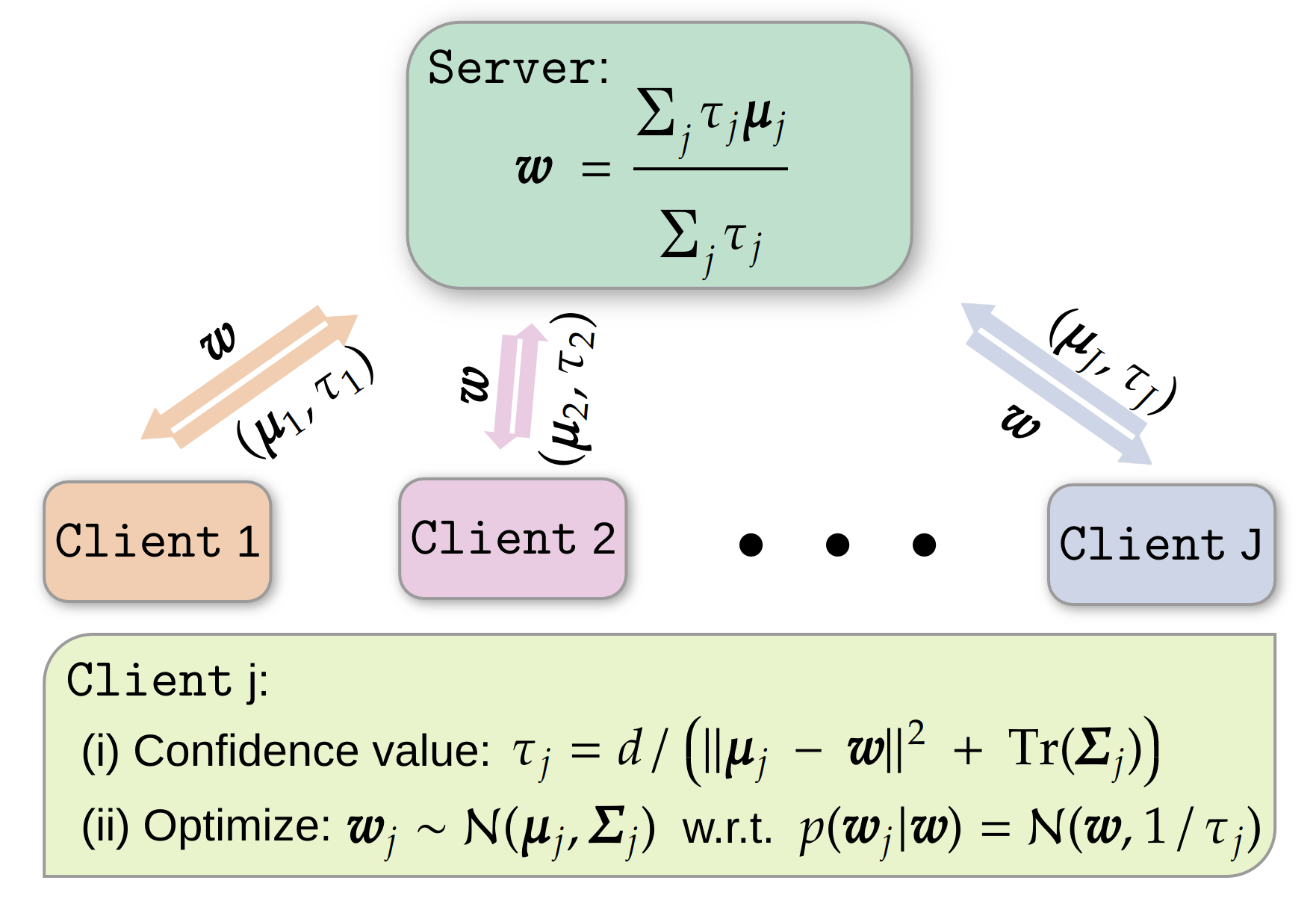}
    \caption{An overview of our  confidence-aware PFL framework.}
    \label{fig:illustration}
\end{figure}

To achieve this, we present a novel Bayesian framework through the lens of a hierarchical latent variable model. In this framework, a latent shared model manages to capture the common trend across local models, and local models adapt to individual environments based on the prior information provided by the shared model. 
In particular, we assume a conditional Gaussian distribution and use variational expectation maximization to optimize the Bayesian models, such that a closed-form solution for the shared model optimization w.r.t.\ clients' confidence values can be derived. The confidence value\footnote{The confidence value is called precision in statistics.} is inversely proportional to the sum of the variance of variational approximation (uncertainty) and squared difference between the mean and the shared model (model deviation). Therefore a low confidence value indicates a high uncertainty or model deviation, and naturally the contribution of this local model to the shared model is considered to be low. An illustration of the proposed confidence-aware PFL framework is presented in \Cref{fig:illustration}.

Additionally, most previous works are solely evaluated under the \textit{label distribution skew} scenario, in this work we also investigate \textit{label concept drift}. As far as we know, there are only few works \cite{marfoq_knnfedper,reddi2021adaptive} considering \textit{label concept drift} in the setting of PFL.
We believe that this scenario is more challenging than \textit{label distribution skew}, since the conditional data distribution from the same label in \textit{label distribution skew} is the same across different clients. In \textit{label concept drift} there can be high discrepancy between local data distributions, indicating higher heterogeneity.

\paragraph{Paper organization.}
Necessary background and notation that will be used in this paper are given in the next section. \Cref{sec:method} presents the theoretical analysis of the proposed framework and the implementation of our algorithm is provided in \Cref{sec:implementation}. Experimental results are presented in \Cref{sec:experiment}. Related works are discussed in \Cref{sec:related_works}. Finally, we conclude in \Cref{sec:conclusion}.
\section{Problem Formulation}
\label{sec:problem-formulation}
In FL a central server orchestrates clients to learn a global objective and the goal is to minimize:
\begin{equation}
    \min_{\bm{w} \in \mathbb{R}^{d} } f(\bm{w} ; \mathcal{D}) := \frac{1}{J} \sum_{j=1}^J f_j(\bm{w} ; \mathcal{D}_j),
\end{equation}
where $J$ is the number of clients, $\mathcal{D}_j$ is the set of data available in client $j$, and $f_j(\bm{w} ; \mathcal{D}_j)$ is the empirical average loss function of the $j$-th client:
\begin{equation}\label{eq:empirical-average-loss}
f_j(\bm{w} ; \mathcal{D}_j) = \frac{1}{n_j} \sum_{i=1}^{n_j} l(\bm{x}_{i}^{(j)}, y_{i}^{(j)}; \bm{w}),
\end{equation}
where $(\bm{x}_{i}^{(j)}, y_{i}^{(j)}) \in \mathcal{D}_j$ is one data point of client $j$, $l(\cdot, \cdot; \bm{w})$ is the loss function using parameters $\bm{w}$ and $n_j := |\mathcal{D}_j|$ is the number of data points on the $j$-th client. Certainly, if $\{D_j\}_1^J$ are non-identically distributed, $\bm{w}$ cannot be optimal for all clients. Instead of using a single global model as in FL, in PFL we aim to solve the composed optimization problem:
\begin{equation}\label{eq:pfl-obj}
\min_{\bm{w}_{1:J} \in \mathbb{R}^{d} } f(\bm{w}_{1:J} ; \mathcal{D}) := \frac{1}{J} \sum_{j=1}^J f_j(\bm{w}_j ; \mathcal{D}_j),
\end{equation}
where $\bm{w}_{1:J}$ is a shorthand for the set of parameters $\{ \bm{w}_1,\cdots,\bm{w}_J \}$ and $\bm{w}_j$ is the personalized parameter for the $j$-th client. 
\section{Confidence-aware Personalized Federated Learning}
\label{sec:method}
In this section, firstly we propose a general Bayesian framework for PFL. Then we derive the optimization methods based on Variational Expectation Maximization (VEM). We therefore name our proposed approach \verb"pFedVEM". We will see that our variational Bayes approach enables the clients to estimate confidence values for their local training results, which automatically adjust the weights in the model aggregation and the strengths of regularization.
\subsection{Hierarchical Bayesian modeling}
To develop a Bayesian framework, we need to obtain a posterior distribution for parameters which we are interested in. Once we have the posterior distribution, any deductive reasoning can be conducted. In the context of PFL, the target is the posterior distribution $p(\bm{w}_j | \mathcal{D}_j)$ of $\bm{w}_j$ for any client $j$.
The most easy way to obtain $p(\bm{w}_j|\mathcal{D}_j)$ is by performing Bayesian inference locally. Given a weak 
prior $p(\bm{w}_j)$, the disadvantage of this approach is the variance of $p(\bm{w}_j | \mathcal{D}_j)$ could be high if the data quantity $|\mathcal{D}_j|$ on client $j$ is limited. However in the context of Bayesian networks, a weak 
prior is almost unavoidable \cite{pmlr-v119-wenzelprior}.

Another way to understand this is, since all clients are running similar tasks, $\mathcal{D}_{\{1,\cdots,J\} \setminus j}$ should be able to provide information to form the posterior of $\bm{w}_j$. In a distributed learning scheme like FL, the $j$-th client has no access to $\mathcal{D}_{\{1,\cdots,J\} \setminus j}$ and it is impossible to obtain the posterior $p(\bm{w}_j | \mathcal{D})$ directly. To overcome this restriction, we introduce a latent variable $\bm{w}$ such that all $\bm w_{1:J}$ depend on $\bm{w}$ and $\bm w$ captures the correlations between different clients. We slightly abuse the notation of $\bm{w}$ which denotes the global model in \Cref{sec:problem-formulation} as the latent variable can also act as a global model fitted on the complete data distribution in our approach.
The relation between $\bm{w}$ and $\bm{w}_{1:J}$ implies conditional independence between clients:
\begin{equation}\label{eq:cond-independence}
    p(\bm{w}_i | \bm{w}) p(\bm{w}_j | \bm{w}) = p(\bm{w}_i,  \bm{w}_j | \bm{w}).
\end{equation}
The conditional distribution $p(\bm{w}_j | \bm{w})$ enables client's models $\bm{w}_{1:J}$ to be specialized in individual environments based on the common trend carried by the latent variable $\bm w$. We now turn to obtain the augmented posterior distribution of $\{ \bm{w},\bm{w}_{1:J} \}$. Using Bayes' rule, the posterior is proportional to the product of the prior and the likelihood function:
\begin{align}
p(\bm{w},\bm{w}_{1:J} | \mathcal{D}) &\propto p(\bm{w}, \bm{w}_{1:J}) p(\mathcal{D} | \bm{w}, \bm{w}_{1:J}) \nonumber  \\
&\overset{\ref{eq:cond-independence}}{=} p(\bm{w}) \prod_{j=1}^J p(\bm{w}_j | \bm{w}) \exp \left( -n_j f_j(\bm{w}_{j} ; \mathcal{D}_j) \right),\nonumber
\end{align}
where $p(\bm{w})$ is the prior distribution of the latent variable, $f_j(\bm{w}_{j}; \mathcal{D}_j) $ is defined in \Cref{eq:empirical-average-loss} and is proportional to the negative of the data log-likelihood on client $j$. 
From the above augmented joint distribution, we see the introduction of the latent variable $\bm{w}$ enables complicated communication across clients. The marginalized joint distribution $p(\bm{w}_{1:J} | \mathcal{D}) = \int p(\bm{w},\bm{w}_{1:J} | \mathcal{D}) d \bm{w}$ can thus be flexible.
\subsection{Variational expectation maximization}
\label{sec:method:vem}
Before an update scheme for $\{ \bm{w},\bm{w}_{1:J} \}$ can be derived, it is necessary to specify the concrete forms for 
the conditional density $p(\bm{w}_j | \bm{w})$.
In this work, we assume an isotropic Gaussian conditional prior $p(\bm{w}_j | \bm{w}) = \mathcal{N}(\bm{w}_j \mid \bm{w}, \rho_j^2 \bm{I})$, where $\rho_j^2$ is the variance of this distribution. A Gaussian conditional implies all clients' parameters are close to this latent variable $\bm{w}$, which is a reasonable assumption since all clients are running similar tasks.
Additionally, this enables a closed form for updating $\bm{w}$. 
In \Cref{sec:experiment}, it will be shown that the isotropic Gaussian assumption works well in practice. 

\textbf{Maximizing the marginal likelihood.} One way to optimize the proposed Bayesian model is Maximum a Posterior Probability (MAP) which seeks a maximizer to the unnormalized posterior, the overall optimization is efficient and easy to implement. However, for MAP the assumption $p(\bm{w}_j | \bm{w}) = \mathcal{N}(\bm{w}_j \mid \bm{w}, \rho_j^2 \bm{I})$ gives rise to a group of hyperparameters $\rho_{1:J}$, which is hard to set. The point estimation of $\bm{w}_j$ can also be unreliable.
To address these issues, we introduce factorized variational approximation $q(\bm{w}_{1:J}) := \prod_{j=1}^J q_j(\bm{w}_j)$ to the true posterior distribution $p(\bm{w}_{1:J} | \mathcal{D})$. In this work, the axis-aligned multivariate Gaussian is used as the variational family, that is, $q_j(\bm{w}_j) = \mathcal{N}(\bm{w}_j \mid \bm{\mu}_j, \bm{\Sigma}_j)$ and $\bm{\Sigma}$ is a diagonal matrix.
To optimize these approximations $\{ q_j(\bm{w}_j) \}_{j=1}^J$, we maximize the evidence lower bound (ELBO) of the marginal likelihood:
\begin{align}\label{eq:elbo}
    &\operatorname{ELBO}\left( q(\bm{w}_{1:J} \right), \rho_{1:J}^2, \bm{w})\\
    &= \sum_{j=1}^J  \mathbb{E}_{q(\bm{w}_j)} [\log p(\mathcal{D}_j | \bm{w}_j)] - \operatorname{KL}[q(\bm{w}_j) \parallel p(\bm{w}_j | \bm{w}, \rho_j^2)].
    \nonumber
\end{align}
The above ELBO can be maximized using VEM through blockwise coordinate descent.
First, to obtain the variational approximations $q(\bm{w}_{1:J})$, for the j-th client we only need to use $\mathcal{D}_j$ and maximize:
\begin{equation}\label{eq:local_objective}
    \mathbb{E}_{q(\bm{w}_j)} [\log p(\mathcal{D}_j | \bm{w}_j)] - \operatorname{KL}[q(\bm{w}_j) \parallel p(\bm{w}_j | \bm{w}, \rho_j^2)]
\end{equation}
Then after these local approximations have been formed, the server attempts to optimize the ELBO in \Cref{eq:elbo} by updating the latent variable $\bm w$ using the client's updated variational parameters. Simplifying \Cref{eq:elbo} w.r.t.\ $\bm{w}$ and $\rho_{1:J}$, we derive the objective function for server:
\begin{align}\label{eq:elbo-server}
\begin{split}
    &\operatorname{ELBO}(\rho_{1:J}^2, \bm{w})\\ 
    &= 
 \sum_{j=1}^J \mathbb{E}_{q(\bm{w}_j)} [\log p(\mathcal{D}_j, \bm{w}_j)] - \mathbb{E}_{q(\bm{w}_j)} [\log q(\bm{w}_j)] \nonumber
\end{split}\\
 &\propto \sum_{j=1}^J \mathbb{E}_{q(\bm{w}_j)} [\log p(\mathcal{D}_j | \bm{w}_j, \bm{w}) + \log p(\bm{w}_j | \bm{w}, \rho_j^2)] \nonumber \\
 &\propto \sum_{j=1}^J \mathbb{E}_{q(\bm{w}_j)} [\log p(\bm{w}_j | \bm{w}, \rho_j^2)].
\end{align}
The last line holds because $\log p(\mathcal{D}_j| \bm{w}_j, \bm{w}) = \log p(\mathcal{D}_j|\bm{w}_j)$ by assumption and $\log p(\mathcal{D}_j | \bm{w}_j)$ does not depend on $\bm{w}$ and $\rho_{1:J}^2$. Setting the first order derivative of \Cref{eq:elbo-server} w.r.t.\  $\bm{w}$ and $\rho_{1:J}^2$ to be zero, we derive the closed-form solutions for these parameters and we define the confidence value $\tau_j := 1/\rho_j^2$:\vspace{+5pt}
\vspace{+3pt}
\fbox{\parbox{0.97\linewidth}{\vspace{-15pt}
\paragraph{Confidence value:}
\begin{equation}\label{eq:confidence}
    \tau_j = d/(\textcolor{Cerulean}{\overbrace{\operatorname{Tr}(\bm{\Sigma}_j)}^{\textit{Uncertainty}}} + \textcolor{Orange}{\underbrace{\| \bm{\mu}_j - \bm{w} \|^2}_{\textit{Model deviation}}}),
\end{equation}
\vspace{-10pt}
}}
\vspace{+5pt}
\fbox{\parbox{0.97\linewidth}{\vspace{-15pt}
\paragraph{Confidence-aware aggregation:}
\begin{equation}\label{eq:server_aggregation}
    \bm{w}^* = \frac{\sum_{j=1}^J \tau_j \bm{\mu}_j }{\sum_{j=1}^J \tau_j};
\end{equation}
\vspace{-10pt}
}}
where $d$ is the dimension of $\bm{w}$, $\operatorname{Tr}(\bm{\Sigma}_j)$ is the trace of the variational variance-covariance parameter, which represents the uncertainty of $\bm{\mu}_j$, and $\| \bm{\mu}_j - \bm{w}\|^2$ represents the model deviation induced by the heterogeneous data distribution. 

\subsection{Advantages} 
From~\Cref{eq:confidence} we see the proposed variational Bayes approach \verb"pFedVEM" enables the clients to estimate the confidence values over their local training results, which involve the derivation of a local model from the global model and the uncertainty of the trained parameters, such that the lower uncertainty and model deviation, the higher confidence. Then in the aggregation (cf.~\Cref{eq:server_aggregation}), the server will form a weighted average from the uploaded parameters w.r.t. the corresponding confidence values. 

Such confidence-aware aggregation has two advantages: \textcolor{Cerulean}{(i) a local model with lower uncertainty has a larger weight. For a similar purpose, many previous works assign weights based on the local data size. However, in the case that a client has a large amount of duplicated or highly correlated data, the uncertainty of our method would be less affected and more accurate.} \textcolor{Orange}{(ii) A local model is weighted less if it highly deviates from the global model. A large distance between the local model and global model indicates that the local data distribution differs a lot from the population distribution. Considering the model deviation will make the aggregation more robust to outliers (e.g.\  clients with data out of the bulk of the population distribution).}

The confidence value also adjusts the regularization effect of the KL divergence term in \Cref{eq:local_objective} during local training. Armed with the isotropic Gaussian assumption, we can now derive the closed form of that KL divergence regularizer. Simplifying w.r.t. $\bm{\mu}_j, \bm{\Sigma}_j$:
\begin{align}\label{eq:closed_kl}
    \begin{split}
        &\operatorname{KL}[q(\bm{w}_j) \parallel p(\bm{w}_j | \bm{w}, \rho_j^2)]\\
        &\propto -\sum_i \log\sigma_{j,i} + (\operatorname{Tr}(\bm{\Sigma}_j) + \| \bm{\mu}_j - \bm{w}\|^2)\tau_j/2
    \end{split}\nonumber\\
    &\geq-\frac{1}{2}\log\operatorname{Tr}(\bm{\Sigma}_j) + \frac{\tau_j}{2}(\operatorname{Tr}(\bm{\Sigma}_j) + \| \bm{\mu}_j - \bm{w}\|^2),
\end{align}
where $\sigma_{j, 1:d}^2$ is the diagonal of $\bm{\Sigma}_j$ and the last line is taken according to Jensen's inequality and convexity of the negative log function.

Based on \Cref{eq:closed_kl}, we observe that the gradient of $\bm{\mu}_j$ w.r.t.\ this KL divergence, i.e.\  $(\bm{\mu}_j - \bm{w})\tau_j$, is rescaled by the confidence value $\tau_j$ such that: \textcolor{Cerulean}{(i) in case a client has a low uncertainty, e.g.\ the local data set is rich, the gradient arising from the log likelihood in~\Cref{eq:local_objective} will be large, while $(\bm{\mu}_j - \bm{w})\tau_j$ is also enlarged due to large $\tau_j$, such that the information of the global model can be conveyed to the local model.}
\textcolor{Orange}{(ii) If $\bm{\mu}_j$ tends to be highly deviated from $\bm{w}$, the regularization effect will not blow up due to reduced $\tau_j$, therefore better personalization can be achieved if data are abundant and highly correlated to the local environments.}
\section{Algorithm Implementation}
\label{sec:implementation}
In this section, we discuss the implementation of our approach, especially the technical difficulties of optimizing~\Cref{eq:local_objective} and present the algorithm of \verb"pFedVEM". 

\textbf{Numerical stability and reparametrization.}
To guarantee the non-negativity of $\bm \Sigma$ and improve the numerical stability, we parameterize the Gaussian variational family of the clients with $(\bm{\mu}_{1:J}, \bm{\pi}_{1:J})$ such that the standard deviation of $q(\bm{w}_j)$ is $\operatorname{diag}(\log(1 + \exp(\bm{\pi}_j)))$. Then in order to conduct gradient descent, we instead sample from the normal distribution and implement for any client j:
\begin{equation}
    q(\bm{w}_j) = \bm{\mu}_j +  \operatorname{diag}(\log(1 + \exp(\bm{\pi}_j))) \cdot \mathcal{N}(0, \bm{I}_d).
\end{equation}

\textbf{Monte-Carlo approximation.} \Cref{eq:local_objective} contains the expectation $\mathbb{E}_{q(\bm{w}_j)} [\log p(\mathcal{D}_j | \bm{w}_j)]$ which rarely has a closed form. We therefore resort to Monte-Carlo (MC) estimation to approximate its value, for $K$ times MC sampling, the objective becomes:
\begin{equation}\label{eq:obj_mc}
    \frac{n_j}{K}\sum_{k=1}^K f_j(\mathcal{D}_j; \bm{w}_{j, k}) - \operatorname{KL}[q(\bm{w}_j) \parallel p(\bm{w}_j | \bm{w}, \rho_j^2)].\nonumber
\end{equation}

\textbf{Head-base architecture.}
Empirically, we find the optimization of $q(\bm w_j)$ is more efficient using a head-base architecture design, which splits the entire network into a base model and a head model. The former outputs a representation of the data and the latter is a linear classifier layer following the base model. Such an architecture is also used in non-Bayesian PFL frameworks~\cite{arivazhagan2019fedper, Collins_icml_fedrep}.
We personalize the head model with \verb"pFedVEM" while letting the base model be trained via \verb"FedAvg". Additionally, the computational demand of \verb"pFedVEM" is thus moderate compared with ~\cite{zhang_pfedbayes}. We do not exclude the possibility that using other federated optimization methods may obtain a better base model, but as we will show in \Cref{sec:experiment}, equipping \verb"pFedVEM" with \verb"FedAvg" already gives significantly improved results, so we leave other combinations for future work. 

Following \cite{dinh_pfedme, zhang_pfedbayes}, at each communication round $t$, the server broadcasts the latent variable $\bm{w}^t$ and base model $\bm{\theta}^t$ to all clients and receives the updated variational parameters and base models from a subset $\mathcal{S}_t$ of clients. The update of server parameters $(\bm{w}, \tau_{1:J})$ depends on each other (cf.\ \Cref{eq:confidence} and \Cref{eq:server_aggregation}), we choose to update $\bm{w}$ first and then $\tau_j$, thus $\tau_j$ can be updated at the client side after receiving the new $\bm{w}$. During the local training, clients first update the head model based on the latest base model and then optimize the base model w.r.t.\ the updated head model. It is worth noting that \verb"pFedVEM" only adds one more scalar ($\tau_j$) to the communication message besides the model parameters, \emph{thus the communication cost is almost unchanged}. We summarize the optimization steps of \verb"pFedVEM" in \Cref{algo:vem}.
\begin{algorithm}[t]
    \caption{pFedVEM: PFL via Variational Expectation Maximization}\label{algo:vem}
    \hspace*{\algorithmicindent} \textbf{Server input:} $T$, $\bm{w}^0$, $\bm{\theta}^0$, $s$ \\
    \hspace*{\algorithmicindent} \textbf{Client input:}
    $\bm{\mu}_{1:J}^{0}$, $\bm{\Sigma}_{1:J}^{0}$,
    $R$, $K$, $\eta$
\begin{algorithmic}[1]
\For{$t=0$ to $T-1$}
\State \textbf{Server executes:}
\For{$j = 1, \dots, J$ \textbf{in parallel} }
\State \textbf{ClientUpdate($\bm{w}^t, \bm{\theta}^t$)}
\EndFor
\State Server selects a random subset of clients $\mathcal{S}_t$ from binomial distribution $B(J, s)$.
\State Each client $j \in \mathcal{S}_t$ sends its updated variational parameters $\bm{\mu}_j^{t+1}$, $\tau_j^{t}$ and base model $\bm{\theta}^{t+1}_j$ to the server.
\LComment{Server optimizes the latent variable}
\State $\bm{w}^{t+1} = \frac{\sum_{j \in \mathcal{S}_t} \tau_j^{t} \bm{\mu}_j^{t+1} }{\sum_{j\in\mathcal{S}_t} \tau_j^{t}}$
\LComment{Server optimizes the base model}
\State $\bm{\theta}^{t+1} = \frac{\sum_{j\in\mathcal{S}_t} n_j\bm{\theta}_j^{t+1}}{\sum_{j\in\mathcal{S}_t}n_j}$
\State \textbf{ClientUpdate($\bm{w}^t, \bm{\theta}^t$)}
\LComment{Client optimizes $\tau_j$}
\State $\tau_j^{t} =d/(\textcolor{Cerulean}{\operatorname{Tr}(\bm{\Sigma}_j^{t+1})} + \textcolor{Orange}{\| \bm{\mu}_j^{t+1} - \bm{w}^{t+1} \|^2})$
\LComment{SGD on $\bm{\mu}_j, \bm{\Sigma}_j$ with $R$ epochs, $K$ sampling times and learning rate $\eta$.}
\State $(\bm{\mu}_j^{t+1}, \bm{\Sigma}_j^{t+1}) \in \arg\min_{(\bm{\mu}_j, \bm{\Sigma}_j)} \mathbb{E}_{q(\bm{w}_j)} [\log p(\mathcal{D}_j | \bm{w}_j)] - \operatorname{KL}[q(\bm{w}_j) \parallel p(\bm{w}_j | \bm{w}^t, 1/\tau_j^t)]$
\LComment{SGD on the base model using the hyperparameters of FedAvg.}
\State $\bm{\theta}^{t+1}_j \in
\arg\min_{\bm{\theta}}\mathbb{E}_{q(\bm{w}_j)}[f(\mathcal{D}_j; \bm{w}_j , \bm{\theta})]$
\EndFor
\end{algorithmic}
\end{algorithm}

\begin{table*}[t!]
\begin{center}
\begin{small}
\begin{tabular}{cccccccc}
\toprule
\multirow{2}{*}{Dataset}     &\multirow{2}{*}{Method} &\multicolumn{2}{c}{50 Clients} 
&\multicolumn{2}{c}{100 Clients}  &\multicolumn{2}{c}{200 Clients}\\
\hhline{~~--||--||--}
    &   &PM     &GM     &PM     &GM     &PM     &GM\\
\midrule
\multirow{11}{*}{FMNIST}
&Local               &$89.2\pm0.1$     &$-$       &$87.5\pm0.1$     &$-$       &$85.7\pm0.1$       &$-$\\
&FedAvg              &$-$        &$83.5\pm0.4$       &$-$     &$85.4\pm0.3$       &$-$       &$85.9\pm0.2$\\    
&FedProx               &$-$      & $84.8 \pm 0.5$       &$-$     &\bftab 86.3 $\pm$ 0.2       &$-$       &\bftab 86.5 $\pm$ 0.1\\
&Scaffold               &$-$      &\bftab 85.6 $\pm$ 0.2       &$-$     &$85.4 \pm 0.1$       &$-$       &$84.6 \pm 0.0$\\
&FedPer               &$91.4\pm0.1$     &$-$       &$90.7\pm0.1$     &$-$       &$89.7\pm0.1$       &$-$\\
&FedRep                 &$91.5\pm0.1$        &$-$       &$90.7\pm0.1$       &$-$       &$89.9\pm0.1$       &$-$\\
&IFCA                 &$84.1\pm1.0$        &$-$       &$85.6\pm0.2$       &$-$       &$86.1\pm0.2$       &$-$\\
&PerFedavg              &$88.7\pm0.2$       &$-$       &$88.6\pm0.1$       &$-$                &$88.3\pm0.2$         &$-$\\
&pFedME             &\bftab 91.9 $\pm$ 0.1       &$82.0\pm0.7$       &\bftab 91.4 $\pm$ 0.1     &$84.4\pm0.6$      &$90.6\pm0.1$       &$85.1\pm0.1$ \\
&pFedBayes             &\bftab 91.9 $\pm$ 0.1       &$83.5\pm0.3$       &$91.3\pm0.1$     &$84.2\pm0.3$      &$90.5\pm0.1$       &$84.4\pm0.1$ \\
&Ours             &$91.8\pm0.1$       &$83.9\pm0.3$       &\bftab 91.4 $\pm$ 0.1     &$85.6\pm0.2$       &\bftab 90.7 $\pm$ 0.1       &$86.2\pm0.2$\\
\midrule
\multirow{11}{*}{CIFAR10}
&Local               &$56.9\pm0.1$     &$-$       &$52.1\pm0.1$     &$-$            &$46.6\pm0.1$       &$-$\\
&FedAvg              &$-$        &$57.7\pm0.9$       &$-$     &$59.4\pm0.6$       &$-$       &$59.2\pm0.3$\\    
&FedProx        &$-$               &$58.0 \pm 0.7$      &$-$       &$59.4\pm0.5$     &$-$       &$59.1\pm0.2$\\
&Scaffold        &$-$               &\bf 60.4 $\pm$ 0.3      &$-$       &$59.8\pm0.2$     &$-$       &$55.4\pm0.3$\\
&FedPer               &$72.7\pm0.3$     &$-$       &$68.4\pm0.4$     &$-$            &$63.4\pm0.3$       &$-$\\
&FedRep                 &$71.4\pm0.3$        &$-$       &$67.4\pm0.4$       &$-$       &$62.8\pm0.2$       &$-$\\
&IFCA                 &$59.4\pm0.8$        &$-$       &$60.1\pm0.5$       &$-$       &$59.5\pm0.5$       &$-$\\
&PerFedavg              &$62.9\pm0.8$       &$-$       &$65.6\pm0.8$       &$-$                &$64.2\pm0.1$         &$-$\\
&pFedME             &$72.3\pm0.1$       &$56.6\pm1.0$       &$71.4\pm0.2$     &\bftab 60.1 $\pm$ 0.3      &$68.5\pm0.2$       &$58.7\pm0.2$ \\
&pFedBayes             &$71.4\pm0.3$       &$52.0\pm1.0$       &$68.5\pm0.3$     &$53.2\pm0.7$      &$64.6\pm0.2$       &$51.4\pm0.3$ \\
&Ours             &\bftab 73.2 $\pm$ 0.2       &$56.0\pm0.4$       &\bftab 71.9 $\pm$ 0.1     &\bftab 60.1 $\pm$ 0.2      &\bftab 70.1 $\pm$ 0.3       &\bftab 59.4 $\pm$ 0.3 \\
\midrule
\multirow{11}{*}{CIFAR100}
&Local               &$34.3\pm0.2$     &$-$       &$27.6\pm0.3$     &$-$            &$22.2\pm0.2$       &$-$\\
&FedAvg              &$-$        &$51.7\pm0.5$       &$-$     &$49.4\pm0.7$       &$-$       &$44.7\pm0.5$\\    
&FedProx               &$-$      &$48.4\pm0.6$       &$-$     &$45.5\pm0.5$       &$-$       &$42.4\pm0.3$\\
&Scaffold               &$-$      &$47.2\pm0.4$       &$-$     &$41.4\pm0.7$       &$-$       &$30.0\pm0.1$\\
&FedPer               &$49.7\pm0.7$     &$-$       &$39.3\pm0.7$     &$-$       &$30.6\pm0.9$       &$-$\\
&FedRep                 &$50.9\pm0.9$        &$-$       &$41.2\pm0.6$       &$-$       &$30.5\pm0.6$       &$-$\\
&IFCA                 &$51.9\pm1.0$        &$-$       &$49.2\pm0.7$       &$-$       &$44.9\pm0.6$       &$-$\\
&PerFedavg              &$52.1\pm0.4$       &$-$       &$48.3\pm0.5$       &$-$                &$40.1\pm0.3$         &$-$\\
&pFedME             &$52.5\pm0.5$       &$47.9\pm0.5$       &$47.6\pm0.5$     &$45.1\pm0.3$      &$41.6\pm1.8$       &$41.5\pm1.6$ \\
&pFedBayes             &$49.6\pm0.3$       &$42.5\pm0.5$       &$46.5\pm0.2$     &$41.3\pm0.3$      &$40.1\pm0.3$       &$37.4\pm0.3$ \\
&Ours             &\bftab 61.0 $\pm$ 0.4       &\bftab 52.8 $\pm$ 0.4       &\bftab 56.2 $\pm$ 0.4     &\bftab 52.3 $\pm$ 0.4       &\bftab 51.1 $\pm$ 0.6       &\bftab 49.2 $\pm$ 0.5\\
\midrule
\multirow{11}{*}{SUN397}
&Local               &$82.4\pm0.9$     &$-$       &$72.0\pm2.2$     &$-$            &$67.4\pm1.4$       &$-$\\
&FedAvg              &$-$        &$73.2\pm0.1$       &$-$     &$72.6\pm0.1$       &$-$       &$72.7\pm0.4$\\    
&FedProx        &$-$               &\bftab 73.7 $\pm$ 0.2      &$-$       &$73.3\pm0.4$     &$-$       &$70.8\pm0.3$\\
&Scaffold        &$-$               & $69.5 \pm 0.4$      &$-$       &$65.5\pm0.4$     &$-$       &$59.9\pm0.6$\\
&FedPer               &$88.4\pm0.4$     &$-$       &$82.3\pm0.2$     &$-$            &$80.0\pm0.1$       &$-$\\
&FedRep                 &$87.8\pm0.3$        &$-$       &$82.1\pm1.2$       &$-$       &$79.6\pm0.4$       &$-$\\
&IFCA                 &$72.5\pm0.5$        &$-$       &$71.5\pm0.5$       &$-$       &$68.0\pm0.5$       &$-$\\
&PerFedavg              &$76.5\pm0.7$       &$-$       &$73.5\pm0.6$       &$-$                &$72.4\pm0.7$         &$-$\\
&pFedME             &$89.6\pm0.7$       &$72.2\pm0.7$       &$82.8\pm2.0$     &$72.3\pm0.6$      &$82.9\pm1.1$       &$73.0\pm1.5$ \\
&pFedBayes             &$83.7\pm0.7$       &$66.1\pm1.0$       &$77.4\pm2.0$     &$65.4\pm0.6$      &$74.6\pm0.3$       &$64.2\pm0.4$ \\
&Ours             &\bftab 91.1 $\pm$ 0.2       &$73.3\pm0.4$       &\bftab 86.6 $\pm$ 1.2     &\bftab 74.1 $\pm$ 0.7      &\bftab 84.5 $\pm$ 0.5       &\bftab 74.3 $\pm$ 0.8 \\
\bottomrule
\end{tabular}
\end{small}
\caption{Average test accuracy of PMs and test accuracy of GM (\% $\pm$ SEM) over 50, 100, 200 clients on FMNIST, CIFAR10, CIFAR100 and SUN397. Best result is in bold.}
\label{tab:acc}
\end{center}
\end{table*}
\begin{figure*}[t!]
    \centering
    \subfloat[\centering Client = 50 \label{fig:cifar100_50}]{\includegraphics[width=0.5\textwidth]{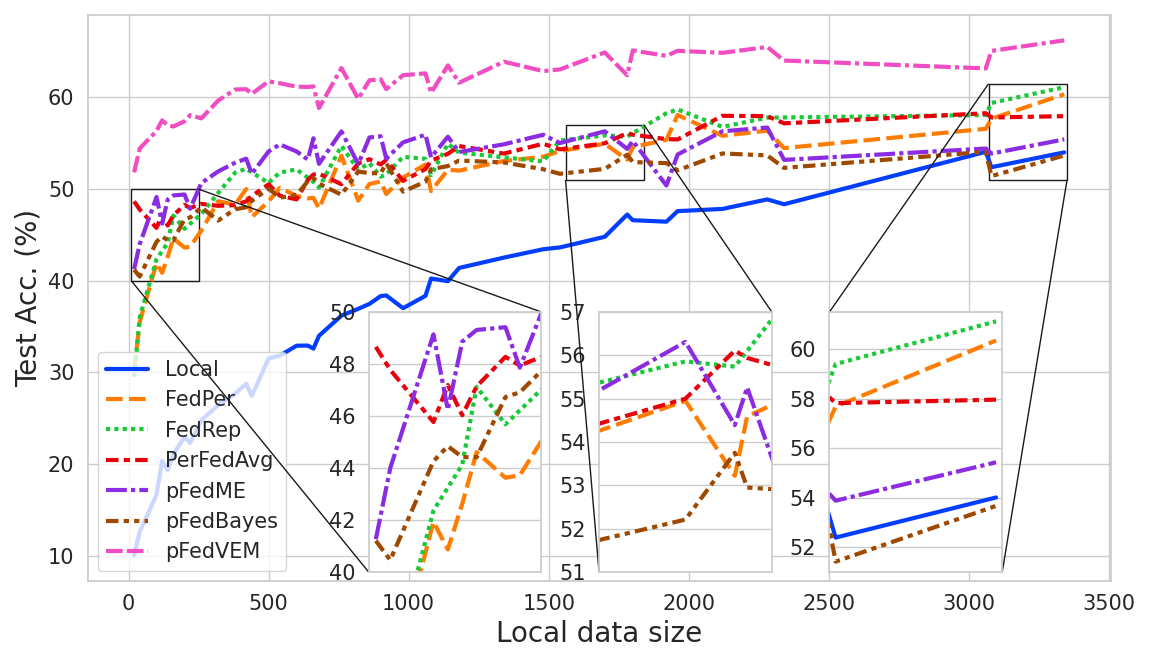}}%
    \subfloat[\centering Client = 100
    \label{fig:cifar100_100}]{\includegraphics[width=0.5\textwidth]{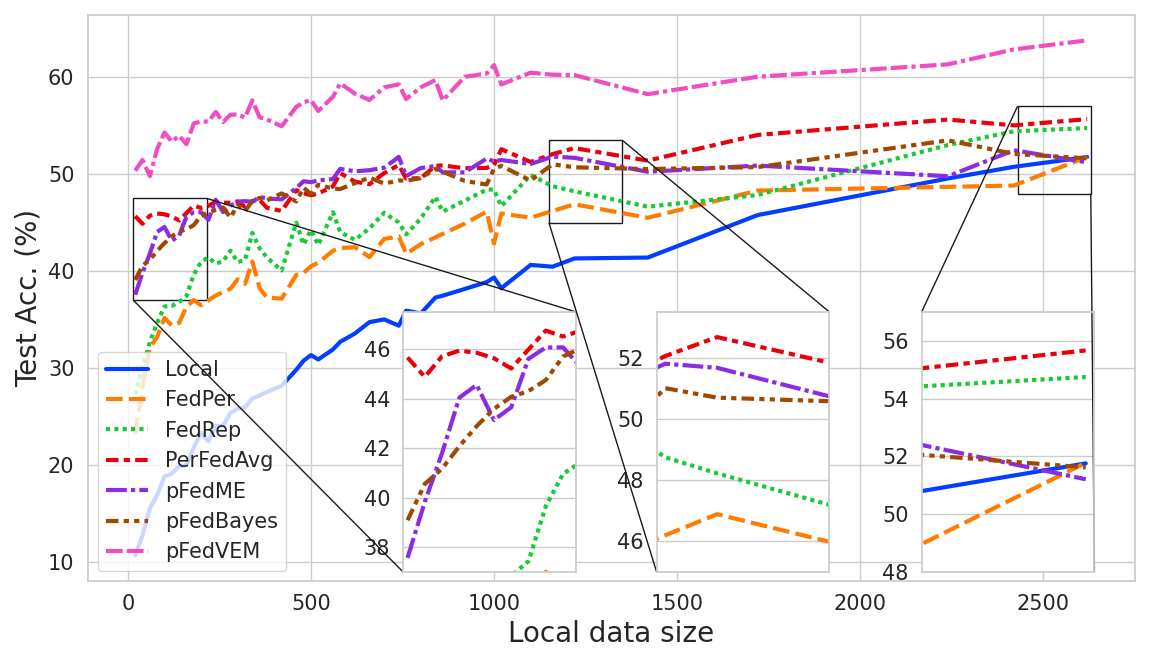}}
    \caption{Test accuracy vs.\ local data size over 50, 100 clients on CIFAR100.}
    \label{fig:acc_vs_size}
\end{figure*}
\section{Experiments}
\label{sec:experiment}
In this section we validate the performance of our approach \verb"pFedVEM" when clients' data are statistically heterogeneous, i.e. \textit{label distribution skew} and  \textit{label concept drift}. We also investigate 
\textit{data quantity disparity} as introduced in~\Cref{sec:intro}. Additionally, we study a case of \textit{feature distribution skew} (a mixture of \textit{label distribution skew} and \textit{label concept drift}), the results are given in \Cref{app:feature_distribution}. We compare our approach with the following FL frameworks: (1) \verb"FedAvg" \cite{pmlr-v54-mcmahan17a}, (2) \verb"FedProx" \cite{MLSYS2020_38af8613}, (3) \verb"Scaffold" \cite{scaffold}, PFL frameworks: (4) \verb"FedPer" \cite{arivazhagan2019fedper}, (5) \verb"FedRep" \cite{Collins_icml_fedrep}, (6) \verb"IFCA" \cite{ifca}, (7) \verb"PerFedAvg" \cite{fallah_perfedavg}, and PFL frameoworks that also output a global model: (8) \verb"pFedME" \cite{dinh_pfedme}, (9) \verb"pFedBayes" \cite{zhang_pfedbayes}, as well as the trivial local training scheme: (10) \verb"Local".
\subsection{Experimental settings}
To evaluate our method, we target image classification problems. Most previous works evaluate on Fashion-MNIST (FMNIST)~\cite{xiao2017online} and CIFAR10~\cite{cifar10} datasets, with which \textit{label distribution skew} can be modeled such that each client has a subset of all labels. We also consider this setting and let each client have 5 out of 10 labels randomly. Furthermore, we model \textit{label concept drift} using CIFAR100~\cite{cifar10} and SUN397~\cite{sun397} datasets. These two hierarchical datasets contain superclasses and subclasses (e.g. subclass couch belongs to the superclass household furniture). We set the classification task to be superclass prediction. CIFAR100 has 20 superclasses and each superclass has 5 subclasses. SUN397 has 3 superclasses and each superclass has 50 subclasses.\footnote{SUN397 dataset is unbalanced so we sample 50 subclasses for each superclass and 100 data points for each subclass. Data are split randomly with 80\% and 20\% for training and testing. Images are resized to 64$\times$64.} 
For each client, we first sample a random subclass from each superclass (1 out of 5 for CIFAR100 and 1 out of 50 for SUN397), then the client's local data is chosen from this subclass, hence \textit{label concept drift} is induced.
To model \textit{data quantity disparity}, we randomly split the training set into partitions of different sizes by uniformly sampling slicing indices, and then distribute one partition to each client. The strength of splitting compared with sampling (without replacement) is that we can use the full training set and local data size can span over a wide range (the sampling range needs to be conservative, otherwise we may run into an out-of-index problem). A concrete example of such a data partition and the resulting local data size distribution is detailed in \Cref{app:datasets}.

When running the experiments the number of communication rounds is set to be 100. We evaluate in all settings the number of clients $J \in \{50, 100, 200\}$; the more clients the more scattered is the training data. To model stragglers, each client has a probability of $0.1$ to send its parameters back to the server at each round of communication. 
Following \cite{zhang_pfedbayes, dinh_pfedme, achituve2021personalized}, we consider a MLP with one hidden layer for FMNIST and LeNet-like CNN \cite{Lecun-lenet} for CIFAR10. The motivation of using small models in FL frameworks is that a single edge device usually has very limited computing power and memory. 
We devise a deeper CNN with 6 layers for CIFAR100 and SUN397 as the tasks on these two datasets are more complex. We illustrate the network architectures in \Cref{app:architecture}.
For experiments on each dataset, we search for hyperparameters with the number of clients $J = 100$ and use these hyperparameters for the other two cases $J = \{50, 200\}$. For \verb"pFedVEM" we search for the learning rate $\eta \in \{0.01, 0.001, 0.0001\}$, initial variance $\rho_{1:J}^2 \in\{1, 0.1, 0.01\}$ and client training epochs $R \in \{5, 10, 20\}$, MC sampling is fixed to 5 times. The optimization method is set to full-batch gradient descent, so we do not need to tune on the batch-size. We extensively search for the baselines' hyperparameters including learning rate, epochs, batch-size and special factors depending on the frameworks. Tables with respective hyperparameters and corresponding searching ranges are presented in \Cref{app:hyperparameters}.

We evaluate both a personalized model (PM) and global model (GM). PMs are evaluated with test data corresponding to the respective labels (for \textit{label distribution skew}) or subclasses (for \textit{label concept drift}) the clients have, while GM is evaluated on the complete test set. All experiments have been repeated for five times using the same group of five random seeds which are used for data generation, parameter initialization and client sampling. We report the mean and its standard error (SEM). All experiments are conducted on a cluster within the same container environment.

\subsection{Results}
\paragraph{Overall performance.}
We first present the average of PMs' test accuracy along with the GM's test accuracy which are two typical evaluation values of PFL and FL frameworks. As shown in \Cref{tab:acc}, \verb"pFedVEM" is competitive on FMNIST and obtains better PMs on CIFAR10 and SUN397, while it's PMs and GM on CIFAR100 significantly outperform the baselines. Based on the model statistics estimated by \verb"pFedVEM" (see \Cref{app:statistics}) we observe that when trained on CIFAR100 local models and confidence values are more scattered, which indicates that \verb"pFedVEM" is more robust and capable at handling high statistical heterogeneity. We also presents the plots of accuracy vs.\ communication rounds to compare the convergence rate of different approaches in \cref{app:convergence_rate}.

Additionally, Bayesian neural networks (BNNs) are known for their exceptional performance when data is scarce~\cite{zhang_pfedbayes}. 
This is because BNNs deal with probability distributions instead of point estimates, and the prior distribution over the weights serves as a natural regularizer. 
So far, we have compared \verb"pFedVEM" to baselines using the full training datasets. To demonstrate the advantage of \verb"pFedVEM" with limited data, we examine two cases: (i) the accuracy of the $10\%$ of clients with the smallest local datasets, (ii) a smaller total number of training samples $|D|$ over all clients. The results are provided in \Cref{app:limited_data}.

\paragraph{Accuracy vs.\ local data quantity.}
FL is a collaboration framework. To attract more clients joining in the collaboration, we need to provide sufficient incentive. Utility gain is a major motivation, which we define as the gap of local model performance between local and federated training. Although average accuracy of PMs in \Cref{tab:acc} reflects the overall utility gain of a federated group, it is deficient in characterizing the utility gain of individual clients, especially considering the utility gain could vary for clients with relatively more or less local data in a federated group. To investigate this, we plot the accuracy over individual data size on CIFAR100 with 50 and 100 clients (see \Cref{fig:acc_vs_size}). Comparing PFL frameworks with \verb"Local", we observe: (1) Generally, clients with relatively fewer data can gain more by joining a collaboration. (2) \verb"FedRep" is good at supporting clients with relatively larger amounts of data, but tends to ignore small clients. (3) \verb"pFedBayes" and \verb"pFedME" can train small clients well, but big clients may not gain or even lose performance by joining these two frameworks, perhaps due to the strong constraint to the global model for a better overall performance. (4) Clients with different data sizes benefit from our confidence-aware PFL framework \verb"pFedVEM".

\paragraph{Ablation study.}
We also conduct ablation studies to understand the two terms \textcolor{Cerulean}{uncertainty} and \textcolor{Orange}{model deviation} in the confidence value of \Cref{eq:confidence} by retaining only the uncertainty term or the model deviation term and evaluate the resulting methods. Based on the results, we see the Uncertainty only method loses more performance when data are non-indentically distributed, indicating that \textcolor{Orange}{model deviation} is helpful when local models tend to deviate from each other (see \Cref{tab:ablation_uncertainty}). The Model deviation only method loses more performance when clients have different data sizes, indicating that \textcolor{Cerulean}{uncertainty} estimation is important under \textit{data quantity disparity} (see \Cref{tab:ablation_model_deviation}). Nevertheless, both ablation methods perform worse than \verb"pFedVEM" under different circumstances.

\begin{table}[t!]
\begin{center}
\begin{small}
\begin{tabular}{cccc}
\toprule
Method     &Hetero.   &Homo.\\
\midrule
pFedVEM             &$61.0\pm0.4$       &$49.4\pm0.2$\\
\textcolor{Cerulean}{Uncertainty}             &$59.9\pm0.2$       &$49.2\pm0.2$\\
\midrule
Mean Diff. (\%)             &$-$\bftab 1.1       &$-0.2$\\
\bottomrule
\end{tabular}
\end{small}
\caption{Test accuracy (\% $\pm$ SEM) of pFedVEM and Uncertainty over 50 clients on CIFAR100. Each client has data of 1 out of 5 subclass (Hetero.) or all 5 subclasses (Homo.) per superclass.}
\label{tab:ablation_uncertainty}
\end{center}
\end{table}
\begin{table}[t!]
\begin{center}
\begin{small}
\begin{tabular}{cccc}
\toprule
Method     &Random   &Equal\\
\midrule
pFedVEM             &$61.0\pm0.4$       &$60.7\pm0.3$\\
\textcolor{Orange}{Model deviation}             &$60.0\pm0.4$       &$60.4\pm0.2$\\
\midrule
Mean Diff. (\%)             &$-$\bftab 1.0       &$-0.3$\\
\bottomrule
\end{tabular}
\end{small}
\caption{Test accuracy (\% $\pm$ SEM) of pFedVEM and Model deviation over 50 clients on CIFAR100. Clients have different local data sizes (Random) or the same amount of local data (Equal).}
\label{tab:ablation_model_deviation}
\end{center}
\end{table}

\section{Related Works}
\label{sec:related_works}
\paragraph{Federated learning.} 
Since the introduction of the first FL framework \verb"FedAvg"~\cite{pmlr-v54-mcmahan17a} which optimizes the global model by weighted averaging client updates that come from local SGD,
many methods \cite{hanzely2021, MLSYS2020_38af8613,wang_obj_inconsistency, nikoloutsopoulos2022} have been proposed to improve it from different perspectives.
\verb"FedNova"~\cite{wang_obj_inconsistency} normalizes the client updates before aggregation to address the objective inconsistency induced by heterogeneous number of updates across clients. \verb"FedProx" \cite{MLSYS2020_38af8613} adds a proximal term to the local objective to alleviate the problem of both systems and statistical heterogeneity. \verb"Scaffold"~\cite{scaffold} uses variance reduction to correct for the client-drift in local updates.
\verb"IFCA"~\cite{ifca} partitioned clients into clusters.
\verb"FedBE"~\cite{chen2021fedbe} and \verb"BNFed"~\cite{yorochikin_bnfed} take the Bayesian inference perspective to make the model aggregation more effective or communication efficient. In particular, a Gaussian distribution is empirically proven to work well on fitting the local model distribution~\cite{chen2021fedbe}. \verb"FedPA" \cite{al-shedivat2021} shows there is an equivalence between the Bayesian inference of the posterior mode and the federated optimization under the uniform prior. \cite{louizos2021} views the federated optimization as a hierarchical latent model and shows that \verb"FedAvg" is a specific instance under this viewpoint. Both works indicate a Bayesian framework is general in modeling federated optimization problems.

\paragraph{Personalized federated learning.} One fundamental challenge in FL is statistical heterogeneity of clients' data \cite{fl_survey}. Many of the FL frameworks described above are developed to prevent the global model from diverging under this problem, while another way to cope with this issue is to learn a personalized model per client \cite{smith_neurips_mocha, arivazhagan2019fedper,mansour2020,dinh_pfedme,shamsian2021,corinzia2021, achituve2021personalized, li2021moon}. \verb"FedPer" \cite{arivazhagan2019fedper} introduces a personalization layer (head model) for each client, while all clients share a base model to learn a representation. \verb"FedLG"~\cite{liang_fedlg} and \verb"FedRep"~\cite{Collins_icml_fedrep} refine such head-base architecture by optimizing the representation learning. Inspired by the Model-Agnostic Meta-Learning (MAML) framework, \verb"PerFedAvg" \cite{fallah_perfedavg} propose to learn a initial shared model such that clients can easily adapt to their local data with a few steps of SGD. \verb"pFedHN" \cite{shamsian2021} use a hypernetwork to generate a set of personalized models. Several works attempt to find the clients with higher correlation and strengthen their collaboration~\cite{zhang2021fomo, smith_neurips_mocha,marfoq_knnfedper, ktpfl_zhang}.  \verb"pFedMe" \cite{dinh_pfedme} introduces bi-level optimization by using the Moreau envelope to regularize a client's loss function. Similar to our work, \verb"pFedBayes" \cite{zhang_pfedbayes} uses Bayesian variational inference and also assumes a Gaussian distribution. However, they develop a framework with a fixed variance for all the local models' prior distribution and therefore do not obtain the personalized confidence value involving the model deviation and uncertainty as \verb"pFedVEM" does.
\section{Conclusion}
\label{sec:conclusion}
In this paper, we addressed the problem of personalized federated learning under different types of heterogeneity, including label distribution skew as well as label concept drift and proposed a general framework for PFL via hierarchical Bayesian modeling. 
To optimize the global model, our method presents a principled way to aggregate the updated local models via variational expectation maximization. 
Our framework optimizes the local model using variational inference and the KL divergence acts as a regularizer to prevent the local model diverge too far away from the global model.
Through extensive experiments under different heterogeneous settings, we show our proposed method \verb"pFedVEM" yields consistently superior performance to main competing frameworks on a range of different datasets.
\subsection*{Acknowledgements}
This research received funding from the Flemish Government (AI Research Program) and the Research Foundation - Flanders (FWO) through project number G0G2921N.


{\small
\bibliographystyle{ieee_fullname}
\bibliography{egbib}
}
\clearpage
\appendix
\paragraph{APPENDIX}
In this appendix we introduce the datasets used in the experiments and clarify our data partition method in \Cref{app:datasets}. Network architectures are illustrated in \Cref{app:architecture}. The grid search of baselines' hyperparameters are detailed in \Cref{app:hyperparameters}. Model statistics on different datasets collected from \verb"pFedVEM" are presented in \Cref{app:statistics}. More experimental results are given in \Cref{app:experiments}.
\section{Datasets}
\label{app:datasets}
In this section we introduce the datasets and data partition method used in this paper. We use Fashion-MNIST (FMNIST) and CIFAR10 to model \textit{label distribution skew}, CIFAR100 and SUN397 to model \textit{label concept drift}.

\paragraph{FMNIST~\cite{xiao2017online}.} This is a dataset of clothing images consisting of 60,000 training data points and 10,000 test data points associated with 10 labels: [T-shirt/top, Trouser, Pullover, Dress, Coat, Sandal, Shirt, Sneaker, Bag, Ankle boot]. 

\paragraph{CIFAR10~\cite{cifar10}.} This dataset consists of 50,000 training data points and 10,000 test data points associated with 10 labels: [Airplane, Automobile, Bird, Cat, Deer, Dog, Frog, Horse, Ship, Truck]. 

\paragraph{CIFAR100~\cite{cifar10}.} This dataset consists of 50,000 training data points and 10,000 test data points associated with 100 labels. The 100 labels (subclasses) are grouped into 20 superclasses such that every superclass contains 5 subclasses, e.g.\ the superclass Household funiture contains [Bed, Chair, Couch, Table, Wardrobe].

\paragraph{SUN397~\cite{sun397}.} This scene category dataset contains 108,753 images from 397 categories. It is arranged in a 3-level tree, we use the first level as superclasses: [Indoor, Manmade outdoor, Nature outdoor] and leaf nodes as subclasses, e.g.\ Indoor contains [Living room, Bedroom, Kitchen, Bathroom, $\dots$]. The number of data points per category and the number of categories per superclass is inconsistent, we take the first 50 categories per superclass and first 100 data points per category according to the ordering in the hierarchy file, then split 80\% and 20\% per category for the training and test set.

Most existing works conduct splitting to generate clients' data. The splitting methods are either subject to a fixed size or a random distribution. The former (e.g. \cite{achituve2021personalized, zhang_pfedbayes}) cannot represent \textit{data quantity disparity}. While the latter methods draw a sequence of fractions per label w.r.t.\ a Dirichlet distribution $Dir_J(\alpha)$ (e.g. \cite{yorochikin_bnfed, fallah_perfedavg}) or a Uniform distribution of a range around 0.5 \cite{shamsian2021} then split and distribute one fraction to each client. As a result the expected local data sizes are still the same across clients, and \textit{data quantity disparity} cannot be well represented. Some work (e.g.\ \cite{dinh_pfedme}) resorts to sampling, but to avoid the out-of-index problem the sampling range is usually conservative.

In this work we conduct splitting by random slicing, the detailed processing steps are as follows:

\textbf{(i) Sampling labels or subclasses:} First we determine what types of data every client contains. Generating a list of labels (for \textit{label distribution skew}) or lists of subclasses (for \textit{label concept drift}), each client samples without replacement from the list of labels or lists of subclasses. When the list is empty, refill and continue until the for-loop over the clients is done.

\textbf{(ii) Sampling data points:} Then we determine the exact data points every client receives. Assume the type of data needs to be partitioned into $M$ parts for $M$ clients, while there are $N$ data points belonging to this type. We first shuffle the data point vector and then draw $M-1$ indices from $1$ to $N-1$ to slice the $N$ data points into $M$ parts. Finally, we distribute the $m$-th part to the $m$-th client allocating this type of data. 

\begin{figure}[h!]
    \centering
    \includegraphics[width=0.6\linewidth]{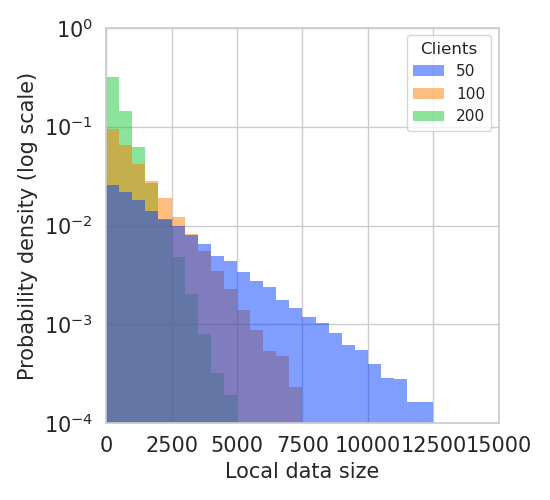}
    \caption{Distribution of local data sizes on FMNIST (setting is consistent with \Cref{tab:acc}). Visualized by 1000 times Monte-Carlo simulation.}
    \label{fig:quantity_fmnist}
\end{figure}

\begin{figure}[h!]
    \centering
    \includegraphics[width=0.6\linewidth]{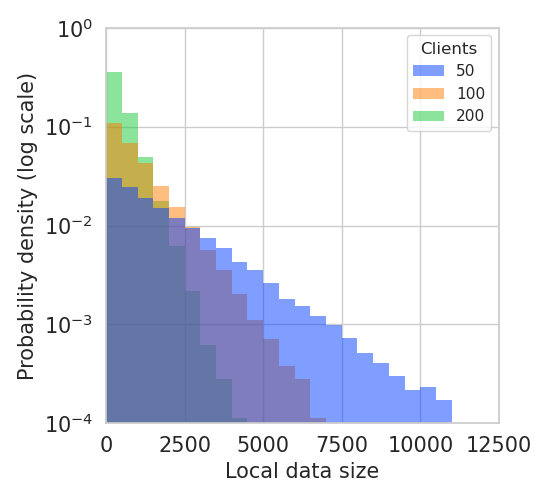}
    \caption{Distribution of local data sizes on CIFAR10 (setting is consistent with \Cref{tab:acc}). Visualized by 1000 times Monte-Carlo simulation.}
    \label{fig:quantity_cifar10}
\end{figure}

\begin{figure}[h!]
    \centering
    \includegraphics[width=0.6\linewidth]{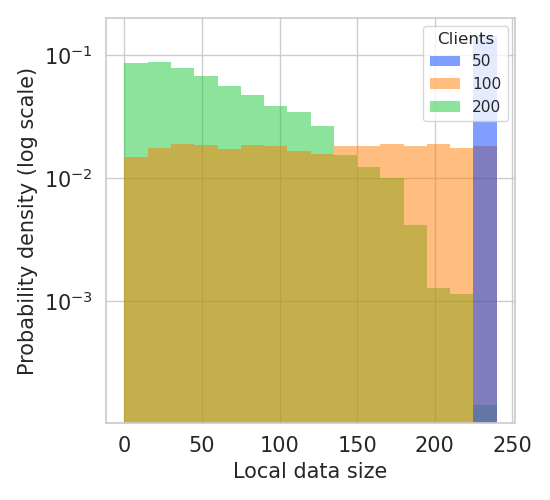}
    \caption{Distribution of local data sizes on SUN397 (setting is consistent with \Cref{tab:acc}). Visualized by 1000 times Monte-Carlo simulation.}
    \label{fig:quantity_sun397}
\end{figure}

\begin{figure}[h!]
    \centering
    \includegraphics[width=0.6\linewidth]{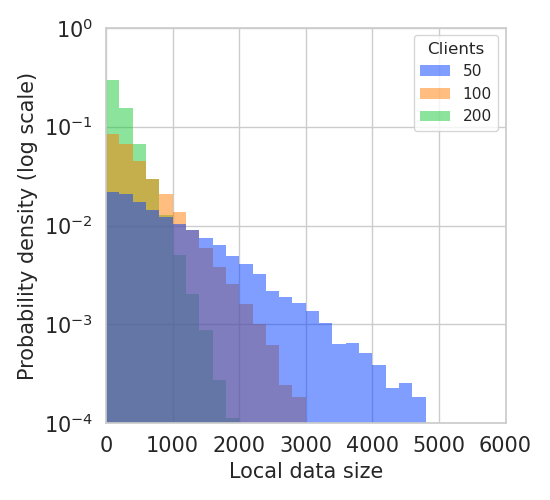}
    \caption{Distribution of local data sizes on CIFAR100 (setting is consistent with \Cref{tab:acc}). Visualized by 1000 times Monte-Carlo simulation.}
    \label{fig:quantity_cifar100}
\end{figure}

There are two strengths of such data splitting. First, the federated group always contains the full training set, while we can make data scatter in different patterns via random seeds. Second, the resulting data partition is close to the negative binomial distribution with one success and thus subject to the nature of \textit{data quantity disparity}, that is, few big datasets are concentrated on few clients, whereas a large amount of data is scattered acros many clients with small dataset sizes. The intuition behind is that when we slice the data points vector at step (ii), $M-1$ indices are uniformly draw from $1,\dots,N-1$. Thus every index approximately conducts a Bernoulli trial with $p=(M-1)/(N-1)$, although in fact they are dependent. We visualize the distribution of local data sizes in \Cref{fig:quantity_fmnist} - \Cref{fig:quantity_cifar100}.
We notice that the distribution on SUN397 is different from others. Recall in our setting, SUN397 has 50 subclasses per subclass. When \#Clients $=50$ every subclass in SUN397 is distributed to exactly one client and when \#Clients $=100$ every subclass is distributed to two clients, thus in these two extreme cases the resulting PDF is either a delta or a uniform function. 
\section{Network Architectures}
\label{app:architecture}
\begin{figure}[h!]
    \centering
    \includegraphics[width=0.6\linewidth]{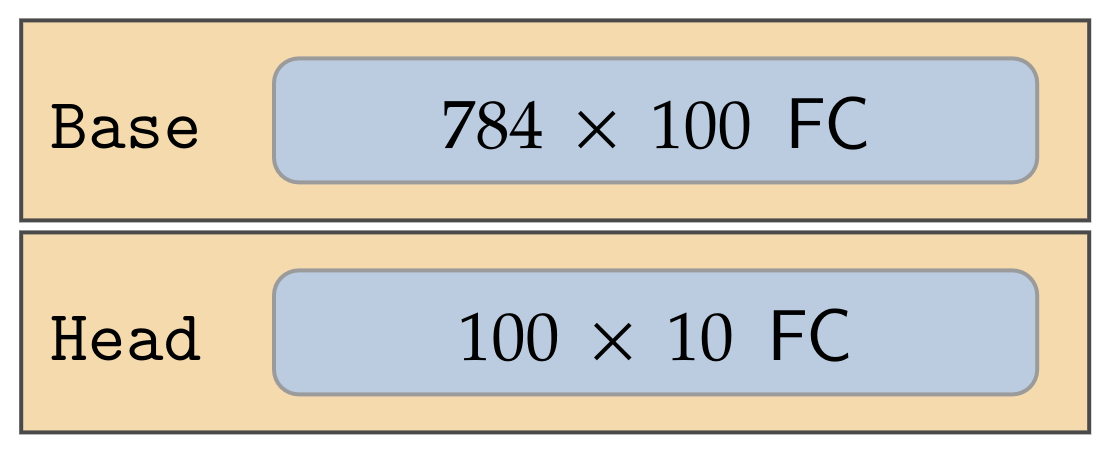}
    \caption{Network architecture for FMNIST.}
    \label{fig:net_fmnist}
\end{figure}

\begin{figure}[h!]
    \centering
    \includegraphics[width=0.6\linewidth]{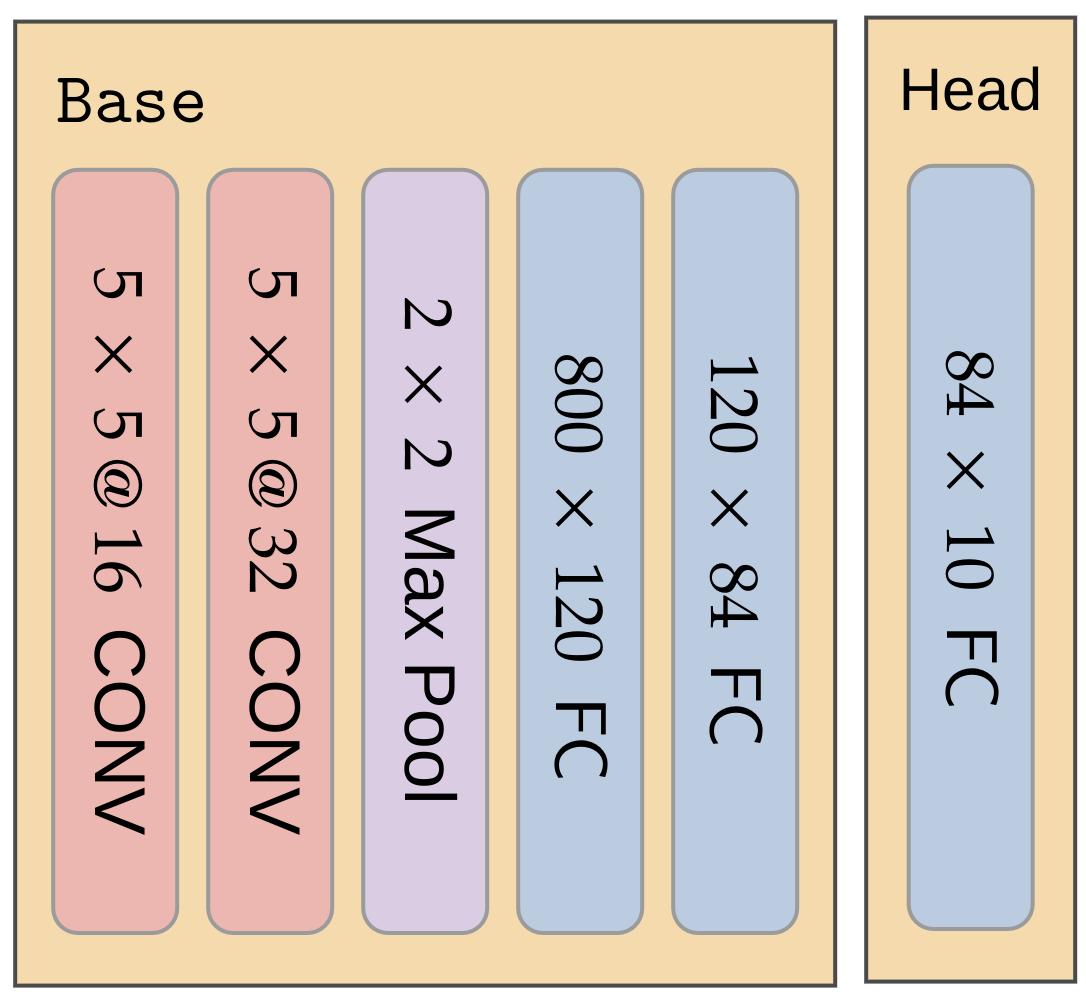}
    \caption{Network architecture for CIFAR10.}
    \label{fig:net_cifar}
\end{figure}

\begin{figure}[h!]
    \centering
    \includegraphics[width=\linewidth]{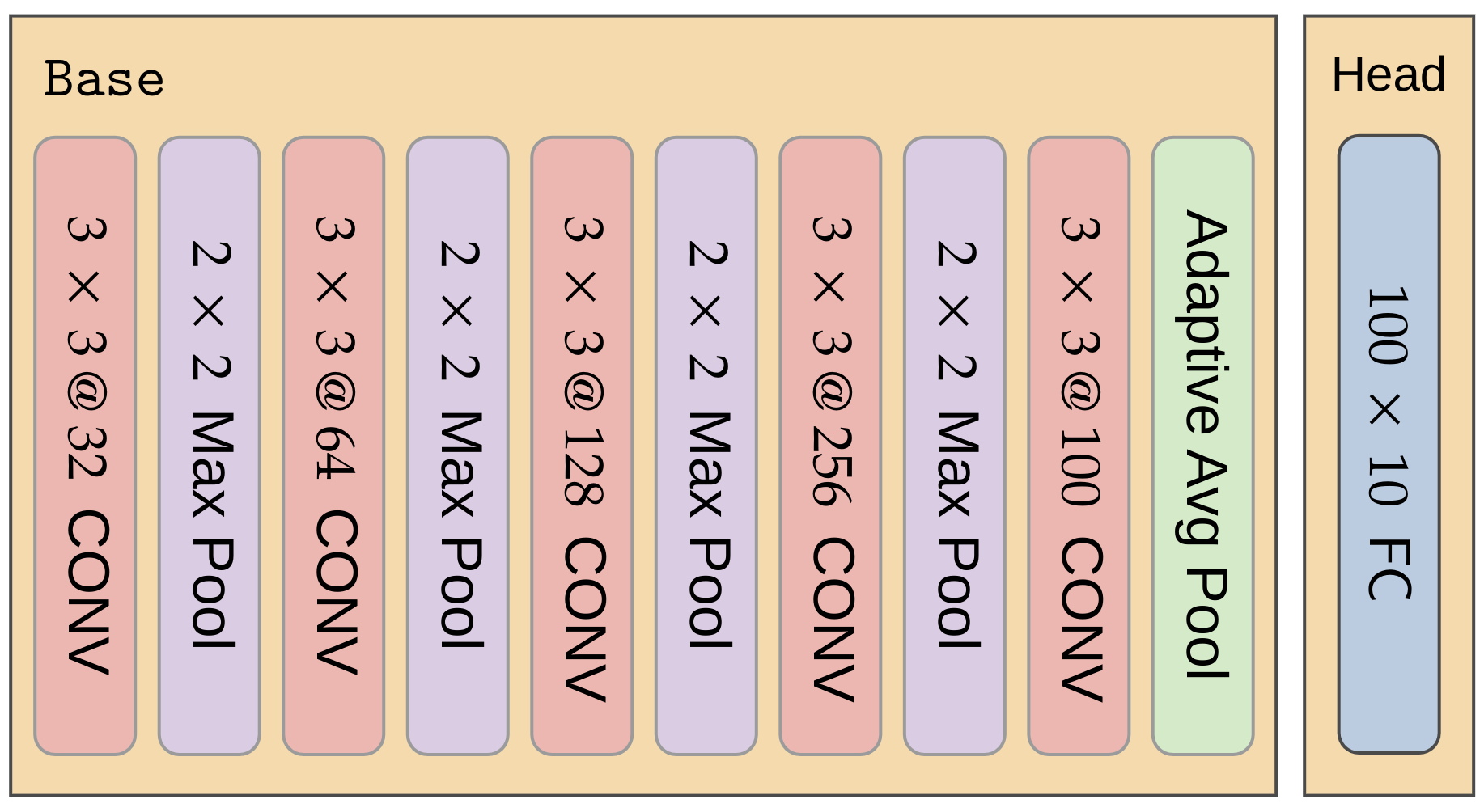}
    \caption{Network architecture for CIFAR100 and SUN397.}
    \label{fig:net_sun397}
\end{figure}

In this section we illustrate the network architectures used in this work, see \Cref{fig:net_fmnist} - \Cref{fig:net_sun397}.
\clearpage
\section{Hyperparameters}
\label{app:hyperparameters}
In this section we describe the hyperparameter search space for each baseline and grid search is used.
The search space is determined according to the best hyperparameters provided in previous works.
The notations of hyperparameters for each framework are detailed below, the search space is summarized in \Cref{tab:hyperparameters}.

\verb"FedAvg": We tune on the learning rate $\eta$, batch-size $B$ and epochs $R$.

\verb"FedProx": We tune on the learning rate $\eta$, batch-size $B$, epochs $R$, and the penalty constant $\mu$ in the proximal term.

\verb"FedPer": We tune on the learning rate $\eta$, batch-size $B$ and epochs $R$.

\verb"Scaffold": We adopt Option 2 and set global learning rate to 1, which are suggested by \cite{scaffold}. We tune on the local learning rate $\eta$, batch-size $B$ and epochs $R$.

\verb"FedRep": We tune on the learning rate $\eta$, batch-size $B$, epochs $R$ for the base model and epochs $K$ for the head model.

\verb"PerFedAvg": We apply the Hessian-free (HF) variant which outperforms the first-order (FO) variant in most settings. We tune on the stepsize $\alpha$ for the adaptation, and the learning rate $\beta$, iterations $R$, batch-size $B$ for the local training of the Meta-model. Before evaluation, every client adapts the Meta-model to the local data with one epoch training using batch-size $B$, which empirically performs better than one-step adaptation.

\verb"IFCA": We follow the configuration for ambiguous cluster structure and define the number of clusters to two, which are suggested by \cite{ifca}. We tune on the learning rate $\eta$, batch-size $B$ and epochs $R$.

\verb"pFedME": Following \cite{dinh_pfedme}, the global model update factor $\beta$ is set to be 1, which is more stable when the number of clients is changed or different random seed is used. We tune on the learning rate $\eta$, batch-size $B$, regularization constant $\lambda$ and local computation rounds $R$.

\verb"pFedBayes": Like \cite{dinh_pfedme}, the global model update factor $\beta$ is set to be 1. Same as \verb"pFedVEM", the batch-size is set to be the local data size. We tune on the learning rate $\eta$, epochs $R$, initial standard deviation $\sigma$ and regularization constant $\lambda$.

\verb"Local": Different from FL frameworks which apply the same hyperparameters across clients, for \verb"Local" we allow every client to train a model locally for 20 epochs while search for the respective best batch-size $B$ and learning rate $\eta$.
\begin{table}[t!]
\begin{center}
\begin{small}
\begin{tabular}{ccc}
\toprule
Method            &Hyperparameter     &Search Range\\
\midrule
\multirow{3}{*}{FedAvg}         &$\eta$       &\{0.01, 0.001, 0.0001\}\\
             &$R$       &\{5, 10, 20\}\\
             &$B$       &\{10, 50, 100\}\\
\midrule
\multirow{4}{*}{FedProx}         &$\eta$       &\{0.01, 0.001, 0.0001\}\\
             &$R$       &\{5, 10, 20\}\\
             &$B$       &\{10, 50, 100\}\\
          &$\mu$       &\{0.1, 1, 10\}\\
\midrule
\multirow{3}{*}{Scaffold}         &$\eta$       &\{0.1, 0.01, 0.001\}\\
             &$R$       &\{5, 10, 20\}\\
             &$B$       &\{10, 50, 100\}\\
\midrule
\multirow{4}{*}{FedPer}         &$\eta$       &\{0.01, 0.001, 0.0001\}\\
             &$R$       &\{5, 10, 20\}\\
             &$B$       &\{10, 50, 100\}\\
\midrule
\multirow{4}{*}{FedRep}         &$\eta$       &\{0.01, 0.001, 0.0001\}\\
             &$R$       &\{5, 10, 20\}\\
          &$K$       &\{5, 10, 20\}\\
             &$B$       &\{10, 50, 100\}\\
\midrule
\multirow{4}{*}{PerFedAvg}         &$\alpha$       &\{0.1, 0.01, 0.001\}\\
&$\beta$       &\{0.1, 0.01, 0.001\}\\
             &$R$       &\{5, 25, 50\}\\
             &$B$       &\{10, 50, 100\}\\
\midrule
\multirow{3}{*}{IFCA}         &$\eta$       &\{0.01, 0.001, 0.0001\}\\
             &$R$       &\{5, 10, 20\}\\
             &$B$       &\{10, 50, 100\}\\
\midrule
\multirow{4}{*}{pFedME}         &$\eta$       &\{0.01, 0.001, 0.0001\}\\
&$\lambda$       &\{1, 10, 15\}\\
             &$R$       &\{5, 10, 20\}\\
             &$B$       &\{10, 50, 100\}\\
\midrule
\multirow{4}{*}{pFedBayes}         &$\eta$       &\{0.01, 0.001, 0.0001\}\\
&$\lambda$       &\{1, 10, 15\}\\
             &$R$       &\{5, 10, 20\}\\
             &$\sigma$       &\{1, 0.1, 0.01, 0.001\}\\
\midrule
\multirow{2}{*}{Local}         &$\eta$       &\{0.001, 0.0001\}\\
             &$B$       &\{10, 50, 100\}\\
\bottomrule
\end{tabular}
\end{small}
\caption{Hyperparameters and the corresponding search space of the baselines.}
\label{tab:hyperparameters}
\end{center}
\end{table}

\section{Model Statistics}
\label{app:statistics}
\begin{figure}[t!]
    \centering
    \subfloat[\centering Communication round = 5 \label{fig:rho_5e}]{\includegraphics[width=0.25\textwidth]{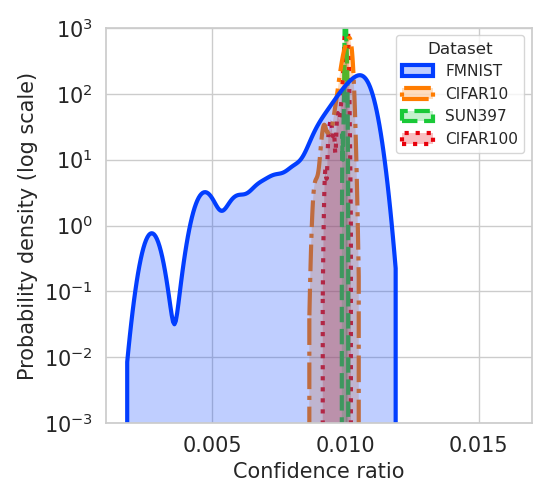}}%
    \subfloat[\centering Communication round = 30
    \label{fig:rho_30e}]{\includegraphics[width=0.25\textwidth]{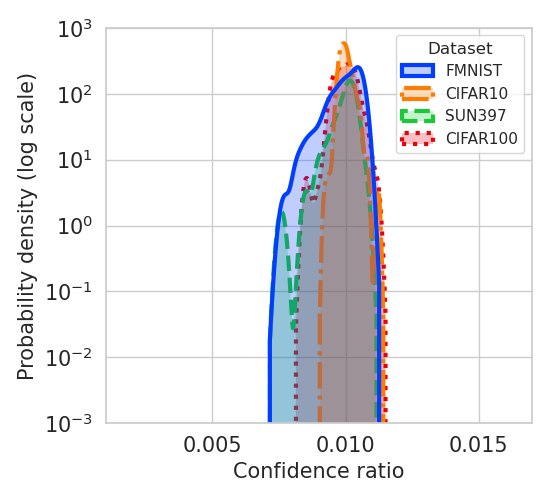}}
    
    \subfloat[\centering Communication round = 60
    \label{fig:rho_60e}]{\includegraphics[width=0.25\textwidth]{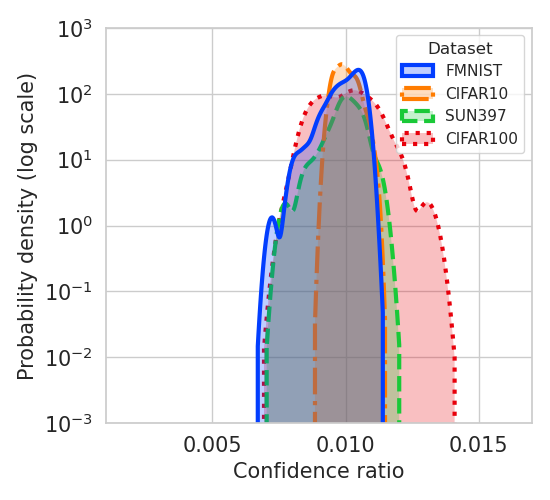}}
    \subfloat[\centering Communication round = 100
    \label{fig:rho_99e}]{\includegraphics[width=0.25\textwidth]{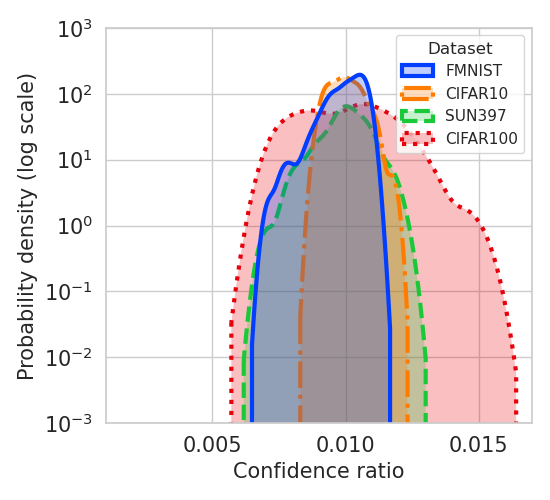}}
    \caption{Distribution of $\tau_{1:J}/ \sum_{j=1}^J\tau_j$ over 100 clients on different datasets. Kernel density estimation is used for the visualization.}
    \label{fig:confidence}
\end{figure}

In this section we discuss the model statistics measured by \verb"pFedVEM". We first investigate the distribution of confidence values in different settings. Since in the model aggregation, the parameters of client $j$ is weighted by $\tau_j/\sum_{j\in\mathcal{S}_t} \tau_j$ (cf.\ \Cref{eq:server_aggregation}), we thus investigate the confidence ratio $\tau_j/\sum_{j=1}^J \tau_j$ instead. 
We collect clients' confidence ratios at different communication rounds from 5 independent runs and visualize the distribution using kernel density estimation with Gaussian kernel.
\Cref{fig:confidence} shows that at the end of training, clients trained on CIFAR100 exhibit the largest variation of confidence ratio. 

\begin{figure}[t!]
    \centering
    \subfloat[\centering Communication round = 5 \label{fig:5e}]{\includegraphics[width=0.25\textwidth]{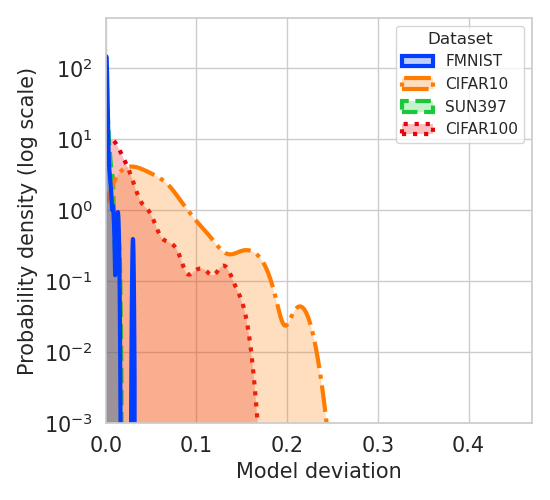}}%
    \subfloat[\centering Communication round = 30
    \label{fig:30e}]{\includegraphics[width=0.25\textwidth]{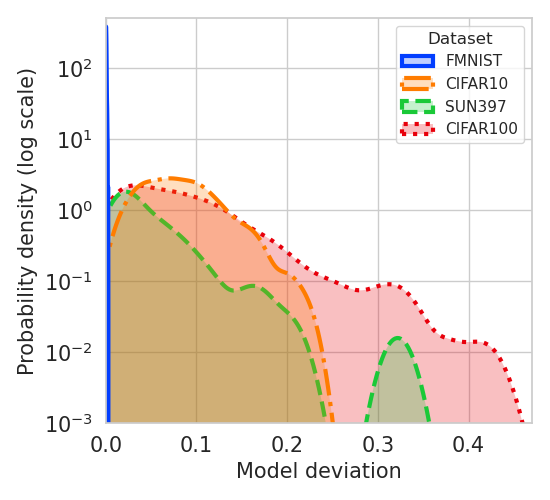}}
    
    \subfloat[\centering Communication round = 60
    \label{fig:60e}]{\includegraphics[width=0.25\textwidth]{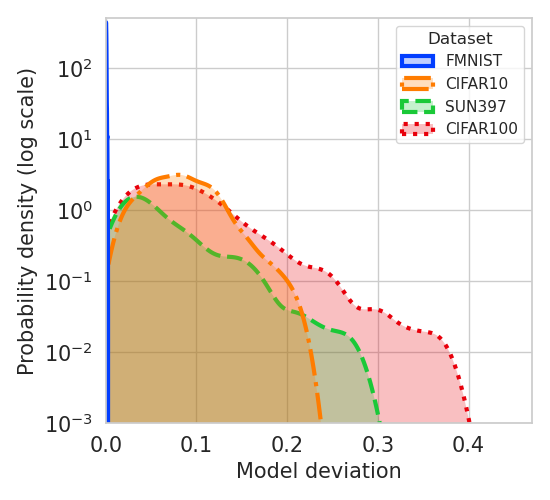}}
    \subfloat[\centering Communication round = 100
    \label{fig:99e}]{\includegraphics[width=0.25\textwidth]{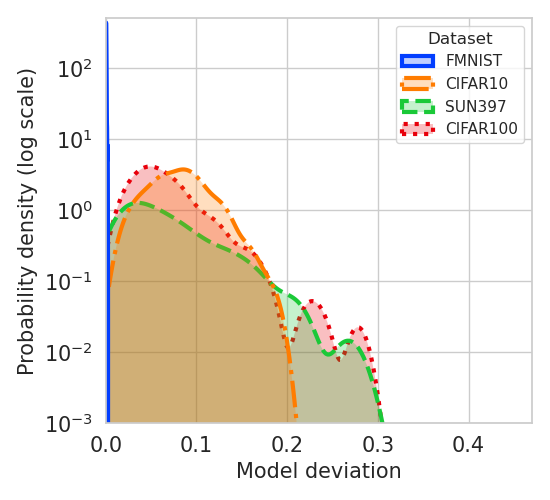}}
    \caption{Distribution of $\|\bm{w}_{1:J} - \bm{w}\|^2/d$ over 100 clients on different datasets. Kernel density estimation is used for the visualization.}
    \label{fig:dist}
\end{figure}

We then investigate the distribution of model deviation. For the $j$-th client the model deviation is defined as $\|\bm{w}_j - \bm{w}\|^2/d$. Similarly, we summarize the results of 5 runs and visualize the distribution using kernel density estimation. \Cref{fig:dist} shows that during training, clients models trained on CIFAR100 spreads out over a larger range. This indicates that the setting of CIFAR100 is highly heterogeneous and this task is a more challenging as the clients' parameters sent back to the server could be severely deviated from each other. In contrast, clients trained on FMNIST concentrates on the global model.
\begin{figure*}[t!]
    \centering
    \subfloat[\centering Client = 50 \label{fig:50}]{\includegraphics[width=0.333\textwidth]{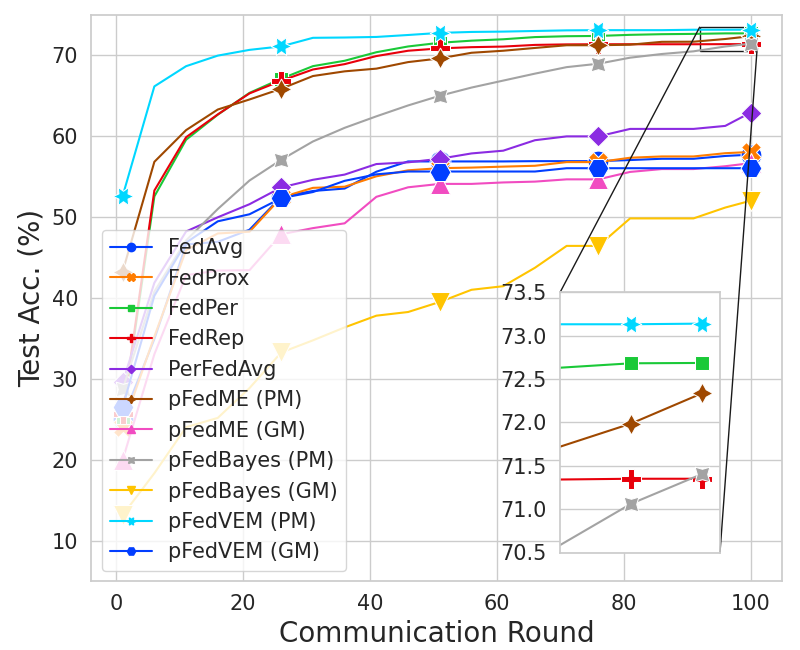}}%
    \subfloat[\centering Client = 100
    \label{fig:100}]{\includegraphics[width=0.333\textwidth]{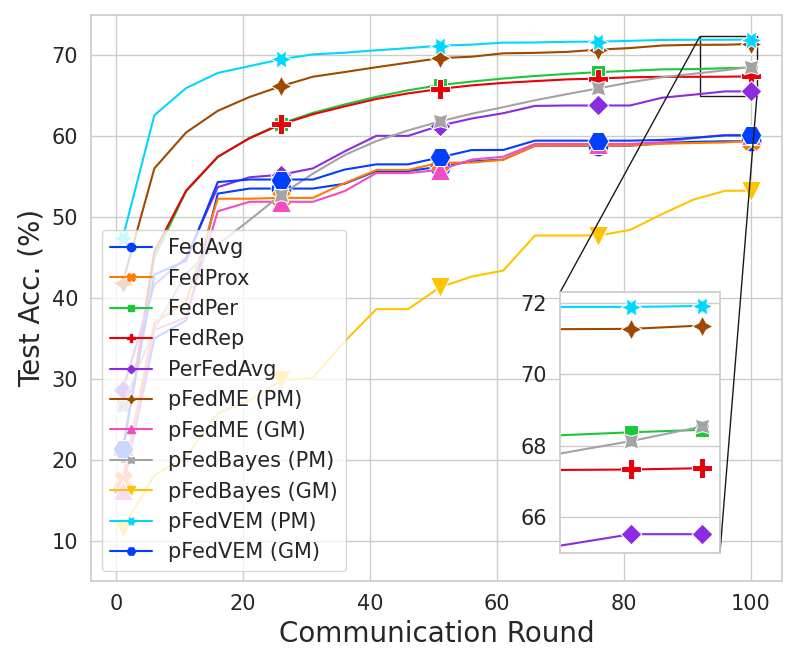}}
    \subfloat[\centering Client = 200
    \label{fig:200}]{\includegraphics[width=0.333\textwidth]{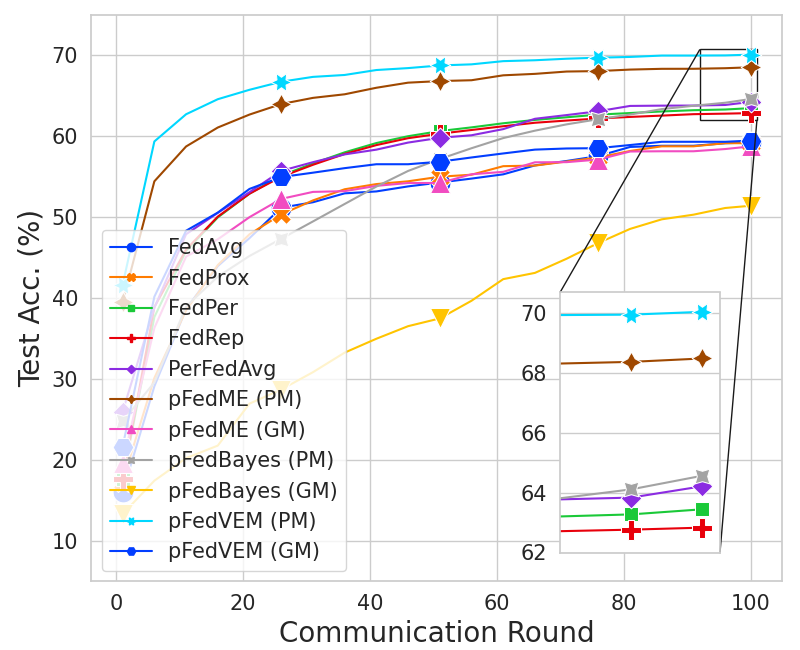}}
    \caption{Convergence rate evaluated by test accuracy vs.\ communication round over 50, 100, 200 clients on CIFAR10.}
    \label{fig:convergence}
\end{figure*}

\section{Additional Experimental Results}
\label{app:experiments}
\subsection{Convergence rate}
\label{app:convergence_rate}
Since the communication bottleneck is one of the main issues of FL, a framework with faster convergence rate is preferred. We therefore present the test accuracy vs.\ communication round plots evaluated on CIFAR10 over 50, 100, 200 clients. As shown in \Cref{fig:convergence}, \verb"pFedVEM" converges faster than other frameworks and already reaches a good accuracy at 30 rounds of communication. This strength of \verb"pFedVEM" is more obvious for 50 clients, where each client is expected to have more data and therefore variational inference performs better.

\begin{table}[t!]
\begin{center}
\begin{small}
\begin{tabular}{cccccc}
\toprule
Dataset  &Method     &Small Clients   &All Clients  &Diff.\\
\midrule
\multirow{6}{*}{\resizebox{0.12\linewidth}{!}{CIFAR10}} &FedPer             &$66.2\pm0.0$       &$68.4\pm0.4$ &-2.2\\
&FedRep             &$64.5\pm0.0$       &$67.4\pm0.4$ &-2.9\\
&PerFedavg             &$51.8\pm0.0$       &$65.6\pm0.8$ &-13.8\\
&pFedME             &$65.0\pm0.0$       &$71.4\pm0.2$ &-6.4\\
&pFedBayes             &$66.9\pm0.0$       &$68.5\pm0.3$ &-1.6\\
&Ours             &\bftab 70.2 $\pm$ 0.0       &\bftab 71.9 $\pm$ 0.1 &-1.7\\
\midrule
\multirow{6}{*}{\resizebox{0.12\linewidth}{!}{CIFAR100}}  
&FedPer             &$30.1\pm0.0$       &$39.3\pm0.7$ &-9.2\\
&FedRep             &$30.8\pm0.0$       &$41.2\pm0.6$ &-10.4\\
&PerFedavg             &$45.2\pm0.0$       &$48.3\pm0.5$ &-3.1\\
&pFedME             &$40.5\pm0.2$       &$47.6\pm0.5$ &-7.1\\
&pFedBayes             &$40.6\pm0.0$       &$46.5\pm0.2$ &-5.9\\
&Ours             &\bftab 50.8 $\pm$ 0.2       &\bftab 56.2 $\pm$ 0.4 &-5.4\\
\bottomrule
\end{tabular}
\end{small}
\caption{Average test accuracy of PMs (\% $\pm$ SEM) of small clients (10\% clients with the smallest local data sizes) and all clients. The difference between the means is shown in the last column. The experimental configuration corresponds to the 100-client setting in \Cref{tab:acc}.}
\label{tab:small_clients}
\end{center}
\end{table}

\begin{table*}[t!]
\begin{center}
\begin{small}
\begin{tabular}{cccccccc}
\toprule
\multirow{2}{*}{Dataset}     &\multirow{2}{*}{Method} &\multicolumn{2}{c}{$|\mathcal{D}|=5000$} 
&\multicolumn{2}{c}{$|\mathcal{D}|=10000$}  &\multicolumn{2}{c}{$|\mathcal{D}|=50000$}\\
\hhline{~~--||--||--}
    &   &PM     &GM     &PM     &GM     &PM     &GM\\
\midrule
\multirow{11}{*}{CIFAR10}
&Local               &$35.7\pm0.3$     &$-$       &$39.6\pm0.1$     &$-$       &$52.1\pm0.1$     &$-$\\
&FedAvg              &$-$        &$41.2\pm0.6$       &$-$     &$47.4\pm1.0$       &$-$     &$59.4\pm0.6$\\    
&FedProx               &$-$      & $41.5 \pm 0.5$       &$-$     &$46.6 \pm 0.5$       &$-$       &$59.4\pm0.5$\\
&Scaffold               &$-$      &$39.7\pm0.6$       &$-$     &$46.4 \pm 0.3$       &$-$       &$59.8\pm0.2$\\
&FedPer               &$40.7\pm0.5$     &$-$       &$51.8\pm0.9$     &$-$       &$68.4\pm0.4$     &$-$\\
&FedRep                 &$40.7\pm0.5$        &$-$       &$51.3\pm0.9$       &$-$       &$67.4\pm0.4$       &$-$\\
&IFCA                 &$41.3\pm0.6$        &$-$       &$47.0\pm0.7$       &$-$       &$60.1\pm0.5$       &$-$\\
&PerFedavg              &$41.4\pm0.4$       &$-$       &$48.2\pm0.3$       &$-$                &$65.6\pm0.8$       &$-$\\
&pFedME             &$44.9 \pm 0.9$       &$36.1\pm1.1$       &$53.1 \pm 1.2$     &$42.5\pm1.8$      &$71.4\pm0.2$     &\bftab 60.1 $\pm$ 0.3 \\
&pFedBayes             &$52.3 \pm 0.5$       &$36.8\pm0.4$       &$58.1\pm0.3$     &$41.9\pm0.9$      &$68.5\pm0.3$     &$53.2\pm0.7$\\
&Ours             &\bftab 57.1 $\pm$ 0.2       &\bftab 45.7 $\pm$ 0.3       &\bftab 61.8 $\pm$ 0.3     &\bftab 50.1 $\pm$ 0.4       &\bftab 71.9 $\pm$ 0.1     &\bftab 60.1 $\pm$ 0.2\\
\bottomrule
\end{tabular}
\end{small}
\caption{Average test accuracy of PMs and test accuracy of GM (\% $\pm$ SEM) over different numbers of training samples of CIFAR10. Other configurations correspond to the 100-client setting in \Cref{tab:acc}. Best result is in bold.}
\label{tab:small_dataset}
\end{center}
\end{table*}

\begin{table*}[t!]
\begin{center}
\begin{small}
\begin{tabular}{cccccccc}
\toprule
\multirow{2}{*}{Dataset}     &\multirow{2}{*}{Method} &\multicolumn{2}{c}{50 Clients} 
&\multicolumn{2}{c}{100 Clients}  &\multicolumn{2}{c}{200 Clients}\\
\hhline{~~--||--||--}
    &   &PM     &GM     &PM     &GM     &PM     &GM\\
\midrule
\multirow{4}{*}{Digit-Five}
&FedProx               &$-$      & $86.2 \pm 0.4$       &$-$     &$86.1 \pm 0.1$       &$-$       & $85.8 \pm 0.3$\\
&pFedME             &$91.2 \pm 0.2$       &$85.6\pm0.4$       &$89.5 \pm 0.3$     &$86.6\pm0.3$      &$88.4 \pm 0.1$     &\bftab 86.9 $\pm$ 0.6 \\
&pFedBayes             &$90.9 \pm 0.2$       &$74.8\pm1.3$       &$88.2\pm0.2$     &$76.1\pm0.6$      &$85.2\pm0.3$     &$74.4\pm0.7$\\
&Ours             &\bftab 92.7 $\pm$ 0.2       &\bftab 86.6 $\pm$ 0.3       &\bftab 91.1 $\pm$ 0.1     &\bftab 87.1 $\pm$ 0.1       &\bftab 89.8 $\pm$ 0.1     &\bftab 86.9 $\pm$ 0.2\\
\bottomrule
\end{tabular}
\end{small}
\caption{Average test accuracy of PMs and test accuracy of GM (\% $\pm$ SEM) over 50, 100, 200 clients on Digit-Five. Best result is in bold.}
\label{tab:digit5}
\end{center}
\end{table*}

\subsection{Limited Data Availability}
\label{app:limited_data}
We examine two cases to demonstrate the advantage of our method \verb"pFedVEM" when data is scarce. First, we show that the small clients, i.e.\ the 10\% of clients with the smallest local data sizes, perform better with \verb"pFedVEM". Based on \Cref{tab:small_clients}, we see the utility gap between the small clients and the overall average is moderate for \verb"pFedBayes" and our method, while our method \verb"pFedVEM" achieves significantly better accuracy than the baselines.
Second, we reduce the number of training samples $|\mathcal{D}|$ that the federated group of all clients has and evaluate the performance of all frameworks. Again, results in \Cref{tab:small_dataset} show that our method performs significantly better than the baselines when data is scarce.

\subsection{Feature Distribution Skew}
\label{app:feature_distribution}
We study a mixed case of \textit{label distribution skew} and \textit{label concept drift}, which is also known as \textit{feature distribution skew}. To this end, we use the Digit-Five dataset consisting of MNIST, SVHN~\cite{svhn}, USPS~\cite{usps}, MNIST-M~\cite{mnistm}, Synthetic Digits~\cite{synthetic_digit} and adopt the following allocation scheme: 1) randomly select a dataset 2) randomly pick 5 labels in the selected dataset 3) randomly select data and distribute to each client.
We tune hyperparameters of three main competing baselines. The results are presented in the \Cref{tab:digit5}. We note that all methods on Digit-Five obtain overall high accuracy, while our method outperforms other methods in this setting of feature distribution skew. 

\end{document}